# Information transmission: Inferring change area from change moment in time series remote sensing images

Jialu Li [a], Chen Wu [a, *], Meiqi Hu [b]

[a] State Key Laboratory of Information Engineering in Surveying, Mapping and Remote Sensing, Wuhan University, Wuhan, 430079, China

[b] School of Geography and Planning, Sun Yat-sen University, Guangzhou, 510275, China

**Abstract:** Time series change detection is a critical task for exploring ecosystem dynamics using time series remote sensing images, because it can simultaneously indicate 'where' and 'when' change occur. While deep learning has shown excellent performance in this domain, it continues to approach change area detection and change moment identification as distinct tasks. Given that change area can be inferred from change moment, we propose a time series change detection network, named CAIM-Net (Change Area Inference from Moment Network), to ensure consistency between change area and change moment results. CAIM-Net infers change area from change moment based on the intrinsic relationship between time series analysis and spatial change detection. The CAIM-Net comprises three key steps: Difference Extraction and Enhancement, Coarse Change Moment Extraction, and Fine Change Moment Extraction and Change Area Inference. In the Difference Extraction and Enhancement, a lightweight encoder with batch dimension stacking is designed to rapidly extract difference features. Subsequently, boundary enhancement convolution is applied to amplify these difference features. In the Coarse Change Moment Extraction, the enhanced difference features from the first step are used to spatiotemporal correlation analysis, and then two distinct methods are employed to determine coarse change moments. In the Fine Change Moment Extraction and Change Area Inference, a multiscale temporal Class Activation Mapping (CAM) module first increases the weight of the change-occurring moment from coarse change moments. Then the weighted change moment is used to infer change area based on the fact that pixels with the change moment must have undergone a change. The experimental results obtained from two global-scale datasets DynamicEarthNet and SpaceNet7 demonstrate that CAIM-Net is superior to the current published methods in speed and accuracy, and has a superior ability for inferring change area from change moment. Moreover, CAIM-Net outperforms representative approaches and achieves Kappa coefficient improvements of 1.12 % and 2.16% for change area detection and of 0.36% and 0.97% for change moment identification. The source code will be available soon at https://github.com/lijialu144/CAIM-Net.

1. Introduction

With the continuous advancement of technology, the Earth's surface cover is experiencing rapid transformations. Factors such as urban expansion (Zhan et al., 2024; C. Zhang et al., 2024), agricultural development (Sun et al., 2024; Qin et al., 2025), and natural disasters (Anno et al., 2024; Wang et al., 2024) are constantly influencing land use patterns. These changes have profound implications for the ecological environment and are closely linked to human survival and development (Chakraborty and Stokes, 2023; T.-H. K. Chen et al., 2023; Xiong et al., 2024). Consequently, monitoring land cover changes (Lou et al., 2024; Xu et al., 2022) has become increasingly important. Remote sensing, as an effective tool for Earth observation, enables us to track these changes with greater accuracy, thereby fostering sustainable development and ensuring the responsible use of our planet's resources

(G. Chen et al., 2024).

Time series remote sensing image change detection (TSCD) entails identifying and locating changes on the Earth's surface by analyzing a series of remote sensing images captured at the same geographic location but at different times (Li and Wu, 2024). This process not only detects the location of these changes but also determines their temporal occurrence, thus facilitating the tracking of land use (Cheng and Li, 2024; Yang et al., 2023), vegetation shifts (Peng et al., 2024; M. Zhang et al., 2024), and natural disasters (Hou et al., 2024; Zheng et al., 2024). Specifically, the detection of change locations can be categorized as the change area detection task, where change area is defined as "pixels representing land cover that has experienced modifications". The determination of when changes occur can be classified as the change moment identification task, where change moment is defined as "identifying between which two images in Time Series Images (TSIs) these changes have occurred" (Li and Wu, 2024). TSCD plays a crucial role in various applications such as Land Use and Land Cover (LULC) monitoring (Masoumi and Van Genderen, 2024; Tong et al., 2025), urban development monitoring (Dong et al., 2024), and environmental monitoring (Liu et al., 2024). In the realm of TSCD, it is common for a pixel to undergo multiple changes. However, detecting the most recent change is particularly important because it not only reflects the current specific status of the object but also provides key information for predicting future trends. By identifying this final change, a more precise assessment of the current situation can be achieved. Consequently, the significance of TSCD cannot be overstated.

In the early stages, traditional methods, such as change vector analysis (CVA) (Bovolo and Bruzzone, 2007; Lian Zhao and Jinshui Zhang, 2011), principal component analysis (PCA) (Deng et al., 2008; Pal et al., 2007), Landsat-based detection of Trends in Distribution and Recovery (LandTrendr) (Kennedy et al., 2010; Runge et al., 2022), and Contiguous Change Detection and Classification (CCDC) (Chen et al., 2021; Zhu and Woodcock, 2014), were commonly used for change detection in remote sensing images. However, these traditional approaches have several notable limitations. They rely predominantly on the temporal dimension, leading to the underutilization of spatial information. Moreover, they can only extract shallow spatial features and thus fail to fully exploit deep spatial features. In contrast, deep learning can automatically learn high-dimensional features that are more representative and significantly enhance feature separability. This capability enables deep learning to excel in handling complex data and patterns. As a result, deep learning has made remarkable advancements in remote sensing image processing, with applications such as change detection (Li et al., 2024; Zheng et al., 2025) experiencing rapid development.

In recent years, driven by the rapid advancement of remote sensing satellites, deep learning has achieved significant progress in Time Series remote sensing images Change Detection (TSCD) tasks. For example, L-UNet (Papadomanolaki et al., 2021) utilized a Siamese UNet structure to extract multiscale spatial features and employed Convolution Long Short-Term Memory (ConvLSTM) as the skip connection to extract difference information at various levels. Building on this, MC$^2$ABNet (Li et al., 2023) further refined spatial and difference feature extraction. Specifically, MC$^2$ABNet also adopted a Siamese UNet structure, incorporating multiscale convolutions and channel attention in each encoding stage to enhance spatial-spectral feature extraction, while applying Convolutional Bidirectional LSTM (ConvBiLSTM) to accurately capture temporal change information for TSCD. Recently, it has been recognized that temporal information is crucial for TSCD, as it enhances the interpretability of changes and aids in identifying and removing outliers, thereby improving model stability and reliability. For instance, UTRNet (Yang et al., 2022) introduced Time-distance-guided LSTM (TLSTM) to effectively handle intra-annual and interannual changes in TSIs. However, the aforementioned

TSCD methods primarily focus on detecting change areas, whereas a crucial aspect of TSCD is identifying the exact moment of change. To address this, Multi-RLD-Net (Li and Wu, 2024) proposed a multi-task network that explored both change area and change moment by effectively utilizing difference features. This network employed LSTMs at different levels to extract difference features across multiple scales, thereby capturing temporal correlations. Despite these advancements, both UTRNet and Multi-RLD-Net are patch-based methods, meaning that spatial information is not fully exploited.

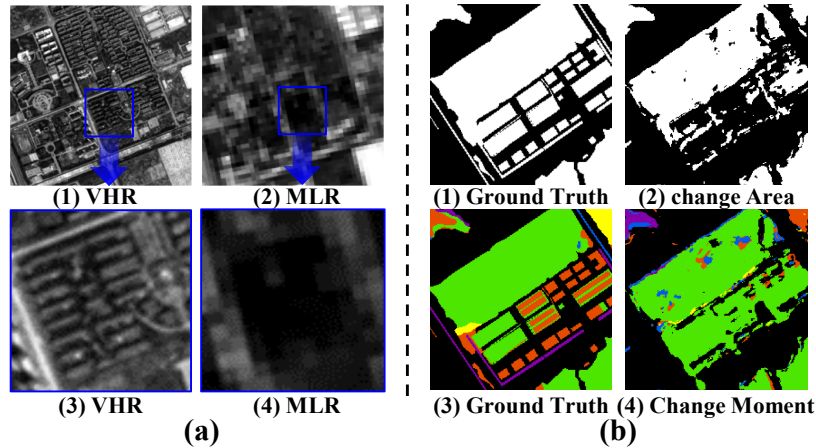

**Fig. 1.** (a) The features of the VHR and MLR images. (b) The change area and change moment results derived from the multitask network. Specifically, (b-1) and (b-2) are ground truth and predicted results for the change area, while (b-3) and (b-4) are ground truth and predicted results for the change moment.

However, current methods face two major challenges. First, images obtained from sensors like Sentinel and Landsat are often used for TSCD because they are more abundant and accessible and can provide TSIs for time series analysis. However, the spatial resolution of these sensor images is usually relatively low, typically falling within the Medium- to Low-Resolution (MLR) range. While VHR images provide clear and distinct object boundaries, medium- and low-resolution TSIs often exhibit ambiguous change boundaries due to the blurred characteristics of objects (Ng and Yau, 2005). A visualization is of this issue is shown in **Fig. 1(a)**. Moreover, the second issue in TSCD is the mismatch between detected change area and change moment, as depicted in **Fig. 1(b)**. Theoretically, region that have undergone change should correspond to a specific change moment, and conversely, pixels with an identified change moment should indicate some form of change. Although multitask networks recognize the association between change area detection and change moment identification, they still treat them as relatively independent tasks in practice, failing to fully exploit their intrinsic close relationship and strong constraints. As a result, the detected change area and change moment do not consistently align.

To jointly address the challenges of unclear change boundaries in TSIs and ensure the consistency between change area detection and change moment identification, this paper proposes a Boundary-Enhanced Change Area and Change Moment Collaborative Detection Network, referred to as CAIM-Net. CAIM-Net comprises three main steps. The first step, Difference Extraction and Enhancement, designs a non-sampling spatial feature extractor preserve the original spatial resolution while extracting features. A boundary enhancement convolution is applied to refine difference features, ensuring that change boundaries are better defined. The purpose of this step is to capture more precise and information difference features from TSIs for subsequent change detection. The second step, Coarse Change Moment Extraction, employs two methods to estimate coarse change moments. The first method extracts change and no-changed features between adjacent images in TSIs and then infers the change moment from these features. The second method concatenates time-series difference features along the temporal dimension, formulating change moment identification as a multi-class semantic segmentation task. This step provides an initial estimation for subsequent fine change detection. The third step, Fine Change Moment Extraction and Change Area Inference, applies Temporal Class Activation Mapping (CAM) at multiple scales to refine the coarse change moment. This technique identifies distinctive features associated with the change along the temporal dimension, enhancing the prominence of specific moment and improving the accuracy of change moment detection. Subsequently, leveraging the inherent property that pixels with an identified change moment must have undergone a change, the fine change area is inferred from the fine change moment. This step aims to concurrently enhance the accuracy of both change area and change moment through their collaborative detection. Finally, the change area and change moment are optimized using Focal Weighted Cross-Entropy Loss (FWCL) to mitigate bias and accumulated errors, yielding improved TSCD results.

The main contributions are summarized as follows:

1) This paper introduces a novel method to infer change moment from change and no-change features between adjacent images in TSIs. Compared to treating change moment identification as a multi-class semantic segmentation task, this approach fully utilizes the information in TSIS, thereby enhancing the reliability of change moment.

2) The change area is inferred from the change moment based on the inherent property that pixels with an identified change moment must have undergone a change. This enables the collaborative detection of change area and change moment, simultaneously improving the accuracy of both tasks.

3) To bridge the interaction gap between change area detection and change moment identification, this paper proposes a boundary-enhanced change area and change moment collaborative detection network, referred to as CAIM-Net. CAIM-Net infers change area from change moment.

4) The proposed CAIM-Net is qualitatively and quantitatively evaluated on two representative TSI datasets, which include change area and change moment labels. Experimental results show that the proposed method improves the Kappa coefficient by about 1.12% or higher over state-of-the-art approaches, even for the more challenging task of change moment identification.

The remainder of this paper is organized as follows. Section 2 provides a detailed elaboration of the proposed CAIM-Net. Section 3 presents the experimental results and analysis. Section 4 provides a discussion of run time and efficiency. Finally, we draw conclusions in Section 5.

## 2. Methodology

This section provides a comprehensive overview of the proposed CAIM-Net. As shown in **Fig. 2**, the overall framework of CAIM-Net comprises three main steps: Difference Extraction and Enhancement, Coarse Change Moment Extraction, and Fine Change Moment Extraction and Change Area Inference. In the first step, given the low spatial resolution of TSIs, we designed a simple and lightweight encoder to extract difference features without downsampling. Initially, ordinary convolutions combined with batch concatenation are utilized to extract spatial features. Subsequently, difference features are computed by taking the absolute value of the difference between adjacent spatial features along the temporal dimension, i.e., $Abs(x_{i+1} - x_i)$. Finally, a boundary enhancement convolution is applied to refine the difference features. In the second step, the boundary enhanced difference features are fed into the spatiotemporal module to extract their spatiotemporal correlation. Two distinct methods are then employed to obtain coarse change moments. This dual approach to extracting coarse change moments not only improves the model's generalization ability but also helps prevent overfitting. In the third step, Temporal CAM is applied at multiple scales to the coarse change moments. This process identifies distinctive features associated with the change along the temporal dimension, resulting in weighted change moments. By summing these weighted change moments across different scales, the fine change moment is obtained. The fine change area is then inferred from the fine change moment, leveraging the inherent principle that pixels with an identified change moment must have undergone a change. This process concurrently enhances the detection accuracy of both the change area and the change moment through their collaborative detection.

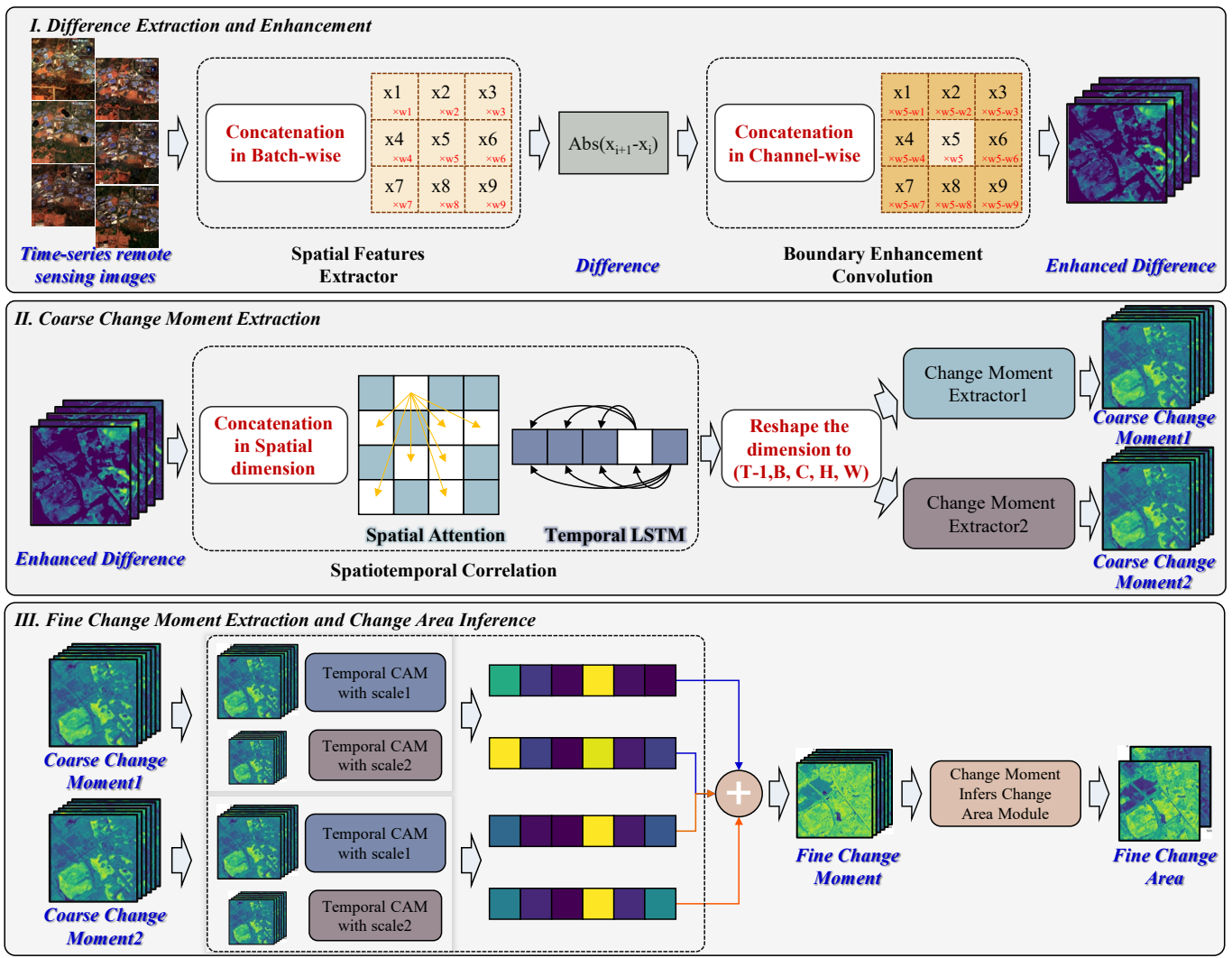

**Fig. 2.** An overview of the proposed CAIM-Net. The network comprises the Difference Extraction and Enhancement, Coarse Change Moment Extraction, and Fine Change Moment Extraction and Change Area Inference.

## 2.1. Difference Extraction and Enhancement

The purpose of Difference Extraction and Enhancement is to capture more refined and effective difference features from TSIs for subsequent change area detection and change moment identification. This process consists of two main components: difference extraction and difference enhancement. In the difference extraction stage, a simple and lightweight encoder is first designed to extract spatial features. The difference features are the obtained by computing the absolute difference between adjacent spatial features along the temporal dimension. In the difference enhancement stage, boundary enhancement convolution is utilized to better distinguish the boundaries between changed and unchanged areas. This technique enhances the edges between different regions in the feature maps, improving the ability to differentiate changed and unchanged regions.

**2.1.1. Difference Extraction**

Change detection typically employs the siamese architecture with an input dimension of $[T, B, C, H, W]$ to extract spatial features through iterative operations. In the proposed CAIM-Net, the time length and batch dimension are concatenated, allowing spatial feature extraction with the combined dimension $[T \times B, C, H, W]$. This approach treats multiple time steps and batches as a single larger batch, simplifies data processing and enhances computational efficiency, which results in faster training and more effective resource utilization.

The encoder consists of two branches, each designed to extract spatial features with different receptive fields. The first branch employs two convolutional layers with $3 \times 3$ kernels, padding 1, and stride 1, to capture fine-grained spatial features. The second branch utilizes two convolutional layers with $1 \times 1$ kernels, padding 0, and stride 1, to extract cross-channel integrated spatial features. The features extracted by these two branches are then summed. This summation not only mitigates spatial information loss but also enhances feature complementarity. The combined features are subsequently passed through a convolutional layer with $3 \times 3$ kernels, padding 1, and stride 1. Up to this point, no downsampling has been applied. The size of the extracted features is $C \times H \times W$, where $C = 64$, and $H = W$ correspond to the input image dimensions. Each convolutional layer is followed by a Group Normalization (GroupNorm) and a Rectified Linear Unit (ReLU) activation function. GroupNorm normalizes each sample independently, which enhances the model's generalization ability by reducing its dependence on batch size. The ReLU activation function introduces non-linearity into the network, enabling it to learn more complex patterns and representations. The output of the encoder can be formulated as follows:

$$E_1 = Conv_1(Conv_1(x)), \qquad (1)$$

$$E_2 = Conv_2(Conv_2(x)), \qquad (2)$$

$$E = Conv_3(E_1 + E_2), \qquad (3)$$

Where $x$ denotes the input TSIs that have undergone batch concatenation, $Conv_1$ represents the convolutional function applied in the first batch, $Conv_2$ represents the convolutional function applied in the second batch, and $Conv_3$ represents the convolutional function applied after summation. $E_1$ is the output of the first branch, $E_2$ is the output of the second branch, and $E$ represents the spatial features extracted by the encoder, which maintain the same spatial resolution as the original input TSIs.

By restoring spatial features from the dimensions of $[T \times B, C, H, W]$ to $[T, B, C, H, W]$, we compute the difference features by subtracting the spatial features of temporally adjacent images along the time dimension and then taking the absolute value. This process extracts the difference features between adjacent images, which can be represented as:

$$D(i) = E(i+1) - E(i), \qquad (4)$$

Where $E$ denotes the input spatial features, $D$ represents the obtained difference features, index $i$ represents the $i-th$ spatial feature of TSIs, where $i = 1, 2, \cdots len(TSIs - 1)$.

The output of difference extraction is the difference features that retain the original spatial size of the input TSIs, without undergoing any downsampling. These difference features serve as the input for the subsequent difference enhancement, where they further facilitate the differentiation between changed and unchanged areas.

**2.1.2. Difference Enhancement**

To obtain more distinct and clear difference features for subsequent change detection, we utilize boundary enhancement convolution to further amplify the difference features, particularly at the boundary between changed and unchanged areas. Boundary enhancement convolution offers several advantages: a robust capability to extract differential information effectively and the incorporation of gradient information encoding capabilities from traditional methods.

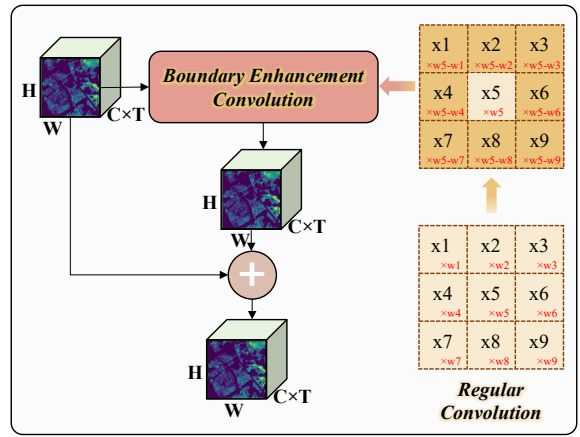

**Fig. 3.** The flowchart of the boundary enhancement convolution used in CAIM-Net.

The primary distinction between regular convolution and boundary enhancement convolution lies in the convolution kernel used in the convolutional layer. The detailed architecture of boundary enhancement convolution is depicted in **Fig. 3** and encompasses four key steps. Firstly, the difference features extracted in the preceding step serve as the input for boundary enhancement convolution. These features are reshaped from $[T - 1, B, C, H, W]$ to $[B, (T - 1) \times C, H, W]$. The rationale behind this reshaping will be elaborated upon in the subsequent text. Secondly, a convolution kernel is constructed specifically for boundary enhancement convolution. The core innovation of boundary enhancement convolution lies in its unique kernel, which differentiates it from regular convolution. Specifically, a $3 \times 3$ matrix is randomly initialized. In regular convolution, this matrix is directly applied to multiply the corresponding pixel values. However, in boundary enhancement convolution, the center pixel is subtracted from surrounding pixels to compute the difference between the center pixel and its surroundings. This operation amplifies the distinguishability between changed and unchanged regions , akin to the principle of Local Binary Pattern (LBP) (Ojala et al., 1994). Thirdly, leveraging the constructed convolution kernel, boundary enhancement convolution extracts boundary information from the difference features. To achieve this, the output channels and the groups of the convolutional layer are set to $(T - 1) \times C$, matching the input feature's channel dimension. Different channels are processed in separate groups, which not only improves computational efficiency but also preserves the temporal order. Lastly, the extracted boundary information is added to the original difference features, yielding the difference features enhanced with boundary details. The final output of the Difference Extraction and Enhancement is the difference features enhanced by boundary information.

**2.2. Coarse Change Moment Extraction**

The primary objective of the Coarse Change Moment Extraction is to provide preliminary results for subsequent fine change detection. In this process, two distinct methods are employed to identify coarse change moments. The first method leverages the change and no-changed features between adjacent images to incrementally deduce the change moment. The second method identifies another coarse change moment by analyzing global features. Together, these methods provide a more comprehensive perspective, enabling the extracted change moments to complement each other. This synergy not only enhances overall accuracy but also improves the robustness of the results.

Before extracting coarse change moments, we designed a spatiotemporal correlation module to capture both spatial and temporal correlations within the time series difference features. In these difference features, regardless of the type of changes that occur, inherent correlations should exist. To establish connections between changed pixels across different regions, we calculate these correlations. Specifically, we employ a Transformer encoder to extract spatial correlations. Before computing the correlations, we reshape the dimensions of the difference features from $[B, T-1, C, H, W]$ to $[B \times H \times W, T-1, C]$. Then, we use a multi-head self-attention mechanism and a feed-forward neural network to compute the spatial correlations within these difference features. The output dimension of the Transformer encoder is $[B \times H \times W, T-1, C]$, where $C = 64$.

To capture temporal correlations, we use an LSTM, which is particularly effective at processing long sequential data. By introducing gating mechanisms, the LSTM mitigates the issues of gradient vanishing and explosion, enabling it to capture long-range dependencies. The input to the LSTM is the difference features with dimensions $[B \times H \times W, T-1, C]$ after spatial correlation computation. To accurately identify the change moment, the hidden state at each time step of the LSTM is outputted, resulting in an output dimension of $[B \times H \times W, T-1, C]$, where $C = 32$. These hidden states are then reshaped to $[T-1, B, C, H, W]$ and fed into two subsequent coarse change moment extractors to extract coarse change moments.

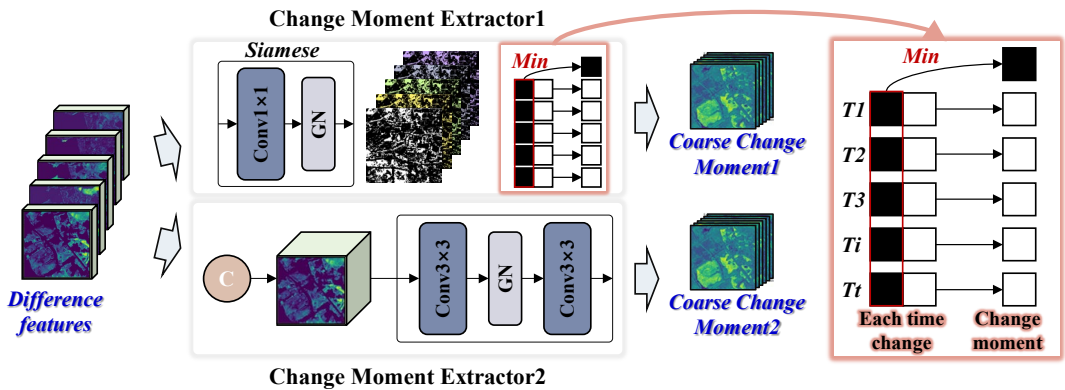

**Fig. 4.** Coarse change moment extractors used in CAIM-Net.

CAIM-Net designed two distinct methods to derive coarse change moment results. The detailed architectures of two methods are illustrated in **Fig. 4**. The first method infers change moment by analyzing change and no-change features between every two adjacent images in TSIs, thereby fully utilizing the available information and enhancing the reliability of change moment inference.

Specifically, we first design a simple siamese network to process the difference features at each time step, obtaining the change and no-change features between every two adjacent images. We apply a simple convolution to compress the dimensionality of the difference features at each time step to 2, where 2 signifies the features of change and no-change at each time step. This convolutional layer is followed by GroupNorm to reduce dependence on batch size. Consequently, we obtained $T-1$ change and no-change features with dimensions $[B, 2, H, W]$.

To infer the change moment, we treat the change feature at each time step as the feature of change at that time step for the change moment, and take the minimum value of the no-change feature across all time steps as the feature of no-change for the change moment, as illustrated in **Fig. 4**. The first coarse change moment extractor yields an output with dimension $[B, T, H, W]$. Thus, in the second dimension $T$, the first value represents the feature of no change, the second value represents the feature of change between the first and second images, and so on, with the last value representing the feature of change between the final image and its preceding image. The output of the first coarse change moment extractor can be formulated as follows:

$$D_i = GN(Conv_4(D)), \qquad (5)$$

$$C_1 = f\{\min(D_i(1)), D_1(2), D_2(2), \cdots, D_i(2)\} \qquad (6)$$

Where $D$ denotes the difference features after spatiotemporal correlation calculation, $Conv_4$ represents the convolutional layer used in the first coarse change moment extractor, and $GN$ represents GroupNorm. $D_i$ denotes the change and no-change features between every two adjacent images, where $D_i(1)$ represents the no-change feature and $D_i(2)$ represents the change feature. $min$ represents the operation of taking the minimum value, $f$ represents the $SoftMax$ activation function, and $C_1$ denotes the output of the first coarse change moment extractor.

The second method treats change moment identification as a multi-class semantic segmentation task, leveraging the analysis of entire feature maps to pinpoint the change moment. Specifically, the temporal and channel dimensions of the obtained difference features are concatenated, transforming the dimensions from $[T-1, B, C, H, W]$ to $[B, (T-1) \times C, H, W]$. Subsequently, two simple convolutional layers are applied to reduce the number of channels to $T$, yielding the second coarse change moment with dimensions $[B, T, H, W]$. These two convolutional layers with $3 \times 3$ kernels, padding 1, and stride 1, the first convolutional layer followed by GroupNorm to mitigate dependence on batch size. The output of the second coarse change moment extractor can be formulated as follows:

$$C_1 = f\left(Conv_5\left(GN(Conv_3(D))\right)\right), \qquad (7)$$

Where $D$ denotes the difference after spatiotemporal correlation calculation, $C_1$ denotes the output of the second coarse change moment, $f$ represents the $SoftMax$ activation function $Conv_4$ represents the convolutional layer used in the second coarse change moment extractor, and $GN$ represents the GroupNorm.

The output of the Coarse Moment Extraction consists of two coarse change moments, extracted through two distinct methods, that serve as preliminary results for subsequent fine change detection.

### 2.3. Fine Change Moment Extraction and Change Area Inference

The primary objective of Fine Change Moment Extraction and Change Area Inference is to achieve the collaborative detection of both change area and change moment. Specifically, by leveraging the inherent property that pixels with a change moment must have experienced a change, the fine change area is inferred from the fine change moment. This inference not only achieves the collaborative detection of the change area and change moment but also simultaneously improves the accuracy of both tasks.

**2.3.1. Fine Change Moment Extraction**

Before inferring the change area, it is necessary to refine the coarse change moment obtained in the previous step to derive the fine change moment. CAM is a powerful technique for interpreting convolutional neural networks and is widely used in weakly supervised tasks. CAM generates a heatmap by combining feature maps from the last convolutional layer with classification weights, effectively highlighting important regions. In the proposed CAIM-Net, we have already identified the coarse change moment. However, our goal is to precisely determine the exact moment when these changes occur. Given its ability to emphasize critical regions, CAM is well-suited for this task, as it enables us accurately identify and weight the exact moment of change. Therefore, CAM is an ideal tool for achieving precise temporal localization of changes.

To effectively utilize CAM in the proposed CAIM-Net, we follow four steps. Firstly, we reshape the dimension of the coarse change moment obtained in the previous step. Specifically, the original dimension $[B,T,H,W]$ is transformed into $[B, 4 \times T, H/2, W/2]$. This transformation is achieved by merging adjacent pixels within an image, enabling temporal CAM to be performed at different scales, thereby enhancing the accuracy of change moment identification. Secondly, for temporal CAM computation, we first reshape the coarse change moment feature from $[B, 4 \times T, H/2, W/2]$ to $[B \times H/2 \times W/2, 4 \times T]$. We then aggregate the coarse change moment features by taking the mean along the temporal dimension. The aggregated features are passed through a fully connected layer to model the linear relationship between the aggregated features and the fine change moment. The output is then processed through a $SoftMax$ function to obtain a probability distribution along the temporal dimension. Thirdly, to highlight the most contributive time points, we apply CAM by performing a matrix multiplication between the coarse change moment features and the weights of the fully connected layer. Each element in the resulting matrix represents the activation score for a specific time point. The CAM scores are then normalized to the range [0,1], emphasizing the most contributive time points while suppressing less important ones. Finally, the normalized CAM scores are rearranged and upsampled to the original spatial resolution via bilinear interpolation. The tensor with the shape of $[B, C, H/2, W/2]$ is upsampled to $[B, C, H, W]$. This step ensures that the CAM scores align with the original spatial dimensions, thereby achieving fine-grained localization of change moment. The calculation of the CAM can be formulated as follows:

$$\bar{C} = f\left(FC(mean(C))\right), \tag{8}$$

$$\overline{CAM} = norm(C \times FC.weight), \tag{9}$$

$$CAM = I(\overline{CAM}), \tag{10}$$

Where $C$ denotes the coarse change moment features after dimension transformation, $mean$ represents the mean operation along the temporal dimension, $FC$ refers to the fully connected layer, $f$ denotes the $SoftMax$ activation function, and $\bar{C}$ represents the probability distribution along the temporal dimension. $\times$ indicates matrix multiplication, $norm$ represents normalization, $\overline{CAM}$ denotes the normalized CAM, $I$ represents bilinear interpolation, and $CAM$ is the final output after upsampling.

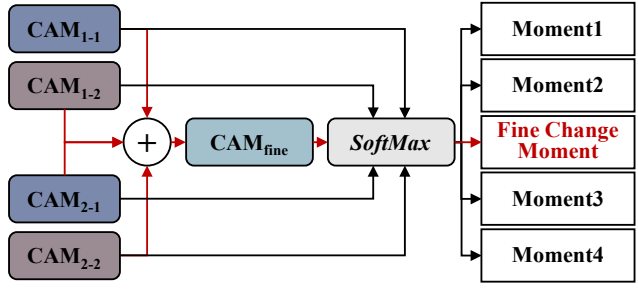

**Fig. 5.** The structure of the fine change moment extractor used in CAIM-Net.

In the preceding CAM computation, it is important to note that the computation is performed at two different scales: $[B, 4 \times T, H/2, W/2]$ and $[B, 16 \times T, H/4, W/4]$. Since the Coarse Change Moment Extraction produces two separate outputs, this results in four CAMs: $CAM_{1-1}$, $CAM_{1-2}$, $CAM_{2-1}$ and $CAM_{2-2}$. These four CAMs are then summed and passed through the $SoftMax$ activation function to generate the Fine Change Moment. Additionally, $CAM_{1-1}$, $CAM_{1-2}$, $CAM_{2-1}$ and $CAM_{2-2}$ are individually processed through $SoftMax$ to produce Moment1, Moment2, Moment3, and Moment4. Moments 1 to 4 serve as supplementary outputs for loss computation. The Fine Change Moment serves as the final outcome for change moment identification and is utilized in both loss calculation and accuracy assessment. The $SoftMax$ activation function can convert the output into a probability distribution, making it is suitable for identifying change moment. In the proposed CAIM-Net, the fine change moment extractor is illustrated in **Fig. 5**. The fine change moment identification can be formulated as follows:

$$Moment1 = f(CAM_{1-1}), \tag{11}$$

$$Moment2 = f(CAM_{1-2}), \tag{12}$$

$$Moment3 = f(CAM_{2-1}), \tag{13}$$

$$Moment4 = f(CAM_{2-2}), \tag{14}$$

$$Fine\ Change\ Moment = f(CAM_{1-1} + CAM_{1-2} + CAM_{2-1} + CAM_{2-2}), \tag{15}$$

Where $CAM_{1-1}$, $CAM_{1-2}$, $CAM_{2-1}$ and $CAM_{2-2}$ represent the input coarse change moments that have undergone multiscale temporal CAM processing, $f$ represents the $SoftMax$ activation function, $Moment1$, $Moment2$, $Moment3$, and $Moment4$ represent supplementary change moment results, while $Fine\ Change\ Moment$ represents the final change moment result.

**2.3.2. Change Area Inference**

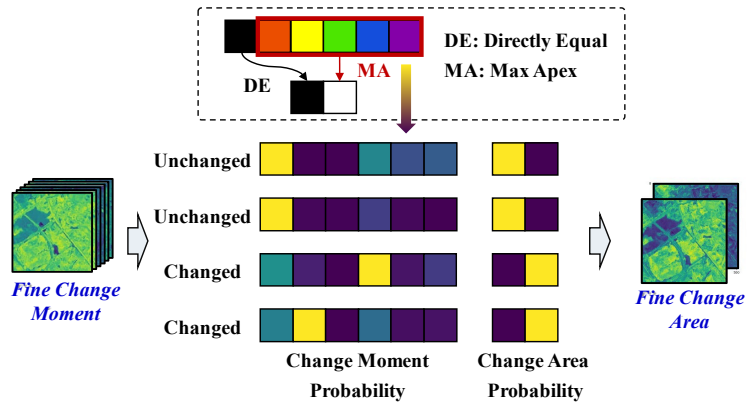

**Fig.6.** The inference utilized for fine change moment to infer fine change area used in CAIM-Net.

To achieve the collaborative detection of change area and change moment, we exploit the inherent property that pixels with an identified change moment must have experienced some form of change. This property allows us to infer the change area directly from the change moment. The inference process is illustrated in **Fig. 6**. Specifically, the no-change probability in the change moment is directly assigned to the no-change probability in the change area. For the change probability, we adopt the maximum change probability detected in the change moment as the change probability in the change area. Subsequently, the $SoftMax$ activation function is applied to normalize the probabilities of change and no-change within the range [0, 1]. For instance, in **Fig. 6**, the top two results depict the inference of unchanged pixels, while the bottom two results depict the inference of changed pixels. The formulation for inferring the change area from the change moment can be formulated as follows:

$$A = f \left( \begin{cases} max(M), & no\ change; \\ M, & change; \end{cases} \right), \tag{16}$$

Where $M$ represents the fine change moment for inferring the change area, $A$ represents the inferred change area, $max$ denotes the operation of selecting the maximum value, and $f$ represents the $SoftMax$ activation function. In the proposed CAIM-Net, change area detection is formulated as a binary classification task distinguishing between change and no change. Therefore, we apply the $SoftMax$ activation function to convert the output into a probability distribution.

**2.4. Loss function**

Cross-entropy loss is widely used in classification tasks due to its effectiveness in measuring the discrepancy between model predictions and ground truth. However, it performs poorly when handling imbalanced datasets. To address this issue, weighted cross-entropy loss introduces class weights to mitigate the impact of sample imbalance, making it particularly useful in tasks where changed and unchanged samples are highly imbalanced.

In addition to class imbalance, remote sensing tasks often suffer an imbalance between hard and easy samples, with easy samples often outnumbering hard ones. This imbalance can lead models to focus predominantly on easy samples while neglecting the more challenging ones (Yang et al., 2022). To mitigate this problem, focal loss has been widely adopted. It not only addressing class imbalance but also effectively reweights samples based on their difficulty, ensuring that hard samples receive greater attention during training.

In addressing both the imbalance among samples, FWCL is employed in the CAIM-Net. The FWCL utilized in change area detection and change moment identification can be formulated as follows:

$$L_f = -\frac{1}{N} \sum_{i=1}^{N} [R(1-\hat{y}_n)^\gamma y_n log\hat{y}_n], \tag{17}$$

The final loss can be formulated as follows:

$$L = L_m + L_A + (L_{m1} + L_{m2} + L_{m3} + L_{m4})/4, \tag{18}$$

In the above equations, $L_f$ represents the FWCL loss used for both change area detection and change moment identification. $L$ represents the overall loss, where $L_m$, $L_A$, $L_{m1}$, $L_{m2}$, $L_{m3}$, and $L_{m4}$ are calculated based on $L_f$. Therefore, the overall loss consists of three components: fine change moment, fine change area, and supplementary change moment. $N$ represents the total number of samples, and $R$ is the ratio of current class samples to all training samples. $\hat{y}_n$ signifies the predicted value, while $y_n$ signifies the corresponding label. $\gamma$ serves as a tunable focusing parameter, operating within the positive domain. In the CAIM-Net, $\gamma$ is also set to 2 for ensuring consistency with the original FWCL (Yang et al., 2022).

## 3. Experiments

### 3.1. Data description and experimental setting

#### 3.1.1. Data description

Due to the limited availability of datasets for TSCD tasks, this paper utilizes two datasets as study areas. The first dataset is the Daily Multi-Spectral Satellite dataset -DynamicEarthNet (Toker et al., 2022), which is available for download from https://mediatum.ub.tum.de/1650201. The second dataset is the SpaceNet 7 Multi-temporal Urban Development Challenge Dataset – SpaceNet7 (Van Etten et al., 2021), which is accessible at https://spacenet.ai/sn7-challenge/. Both the DynamicEarthNet and SpaceNet7 datasets are employed to assess the performance of proposed CAIM-Net.

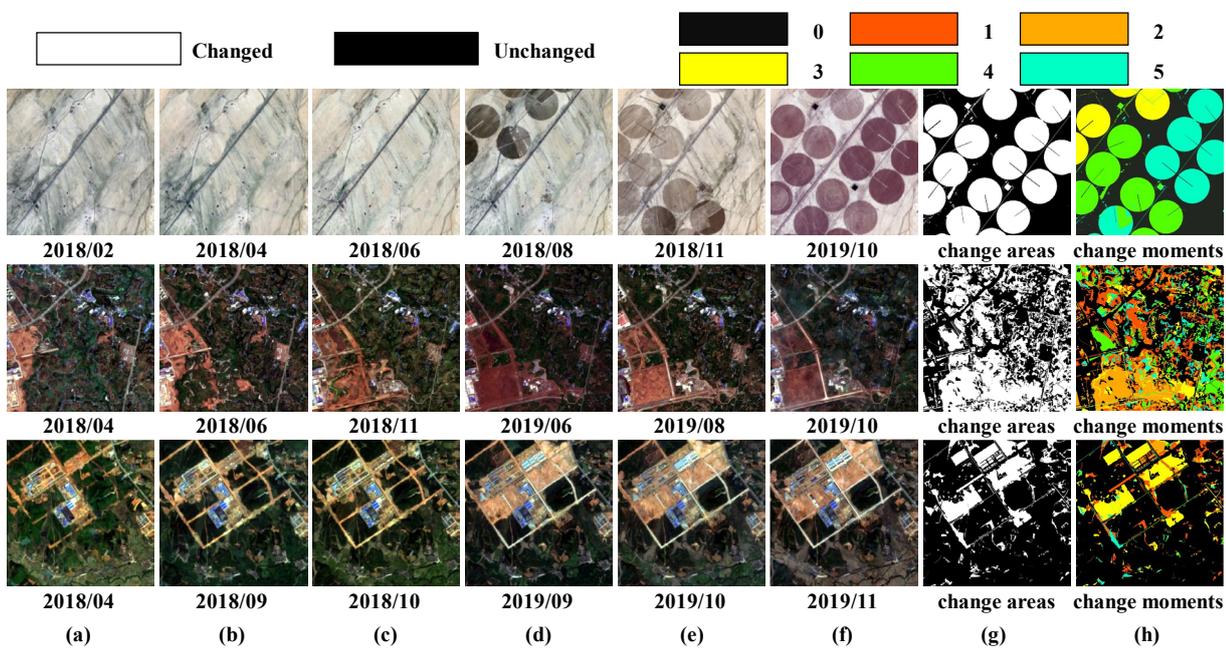

**Fig. 7**. Illustration of TSIs applied to various scenes in the DynamicEarthNet dataset. The TSIs are denoted from (a) to (f). The (g) column depicts change area labels, with changes indicated in white and unchanged pixels in black. The (h) column depicts change moment labels, with different colors representing different change moments.

1) DynamicEarthNet: The DynamicEarthNet dataset consists of monthly Sentinel-2 (S2) images captured between January 2018 and December 2019. It includes thirteen spectral bands that span a wide range of wavelengths, from visible to short-wave infrared. These bands are: Coastal Aerosol (B1), Blue (B2), Green (B3), Red (B4), Vegetation Red Edge (B5, B6, B7, B8A), Near-Infrared (B8), Water Vapor (B9), SWIR-Cirrus (B10), and SWIR (B11, B12). Each band has different spatial resolutions of 10 m, 20 m, and 60 m. For model training, we specifically selected four bands with a 10 m resolution: Blue, Green, Red, and Near-Infrared. The dataset encompasses 75 Areas of Interest (AOIs) distributed globally, each exhibiting diverse land cover changes, including impervious surface, agriculture, vegetation, wetlands, soil, water, and snow. Each AOI contains a set of 24 Sentinel-2 images, capturing seasonal variations, with an image size of $1024 \times 1024$ pixels. The total coverage area of the DynamicEarthNet dataset is 188,743km² km². Representative images, along with their corresponding change areas and change moments, are illustrated **in Fig. 7**.

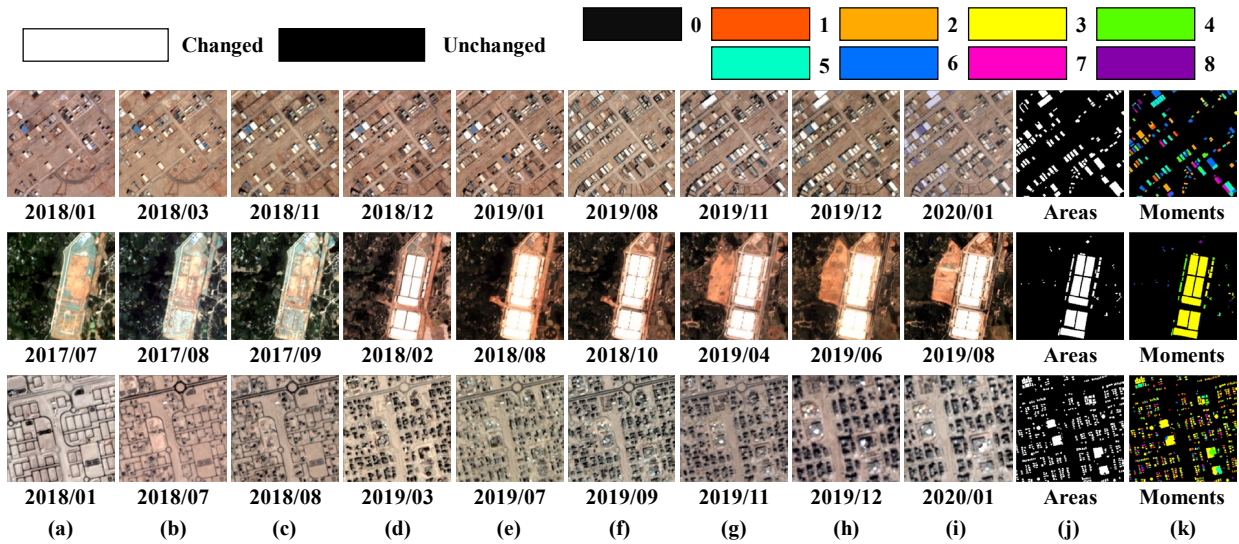

**Fig. 8**. Illustration of TSIs applied to various scenes in the SpaceNet7 dataset. The TSIs are denoted from (a) to (i). The (j) column depicts change area labels, with changes indicated in white and unchanged pixels in black. The (k) column depicts change moment labels, with different colors representing different change moments.

2) SpaceNet7: The SpaceNet7 dataset was released to support multitemporal building detection in the NeurIPS 2020 challenges. It comprises monthly Planet satellite images collected from July 2017 to January 2020, covering a period of over two years. The dataset includes four spectral bands—Red, Green, Blue, and Near-Infrared—each with a spatial resolution of 4 m. We utilized all four bands to train the model. The SpaceNet7 dataset encompasses 100 AOIs worldwide, each characterized by significant changes in building structures. For each AOI, approximately 24 Planet images were selected to capture seasonal variations. These images have a size of approximately $1024 \times 1024$ pixels. The total coverage area of the SpaceNet7 dataset reaches 41,000km². **Fig. 8** illustrates representative images from the SpaceNet7 dataset, along with their corresponding change areas and change moments. We evaluate the performance of our proposed CAIM-Net using both the DynamicEarthNet and SpaceNet7 datasets. The process for obtaining change area and change moment labels in the DynamicEarthNet and SpaceNet7 datasets is outlined below.

In the DynamicEarthNet dataset, each image within AOIs is annotated with pixel-wise LULC semantic labels, while in the SpaceNet7 dataset, each image is labeled with pixel-wise building semantic labels. These semantic labels serve as ground truth for identifying land cover and building changes over the observation period. Specifically, for each AOI with time series semantic labels, we first compute the difference between every pair of adjacent semantic labels to generate change labels for consecutive images. Based on these change labels, we derive the corresponding change area and change moment labels. For the change area label, a pixel is marked as 'unchanged' if it remains unchanged across all change labels; otherwise, it is marked as 'changed'. For the change moment label, a pixel is marked as 'unchanged' if it remains unchanged across all change labels; otherwise, the moment of change is recorded. If a pixel undergoes multiple changes, the last change moment is recorded as the change moment label. As illustrated in **Fig. 7** and **Fig. 8**, the labeling scheme is as follows: '0' indicates unchanged, '1' indicates change occurring between the first and second images, '2' indicates change occurring between the second and third images, and so on. This process of obtaining change area and change moment labels in CAIM-Net is consistent with that in Multi-RLD-Net (Li and Wu, 2024).

**3.1.2. Experimental setting**

1) Experimentation Details: The proposed CAIM-Net framework is implemented on the PyTorch platform using an NVIDIA GeForce RTX 3090Ti GPU. We employ the Adaptive moment estimation (Adam) optimizer (Kingma and Ba, 2017) with an initial learning rate of $1e^{-4}$. Our CAIM-Net framework achieves marked efficiency improvements using fewer epochs. Specifically, for the DynamicEarthNet dataset, the training epoch is set to 100, while for the SpaceNet7 dataset, it is set to 50. The input TSIs image size is $64 \times 64$, and the batch size is set to 96 for DynamicEarthNet dataset and 64 for SpaceNet7 dataset. The DynamicEarthNet dataset contains numerous images affected by clouds cover, as cloud conditions were not considered during image acquisition. These cloudy images were removed during the training. Finally, we selected 30 TSIs cubes from 55 labeled AOIs, each containing six images. The training, validation, and testing data were split in an 8:1:1 ratio, consistent with the setups in RLD-Net and Multi-RLD-Net. In total, approximately 36,864 images were used for training, 4,608 images for validation, and 4,608 images for testing. The SpaceNet7 dataset comprises 60 TSIs cubes with building semantic labels. While the SpaceNet7 dataset is also affected by cloud cover, its impact is less severe than in DynamicEarthNet dataset. Cloudy images were similarly removed during training. Finally, 60 TSIs cubes each containing nine images from different time periods. The training, validation, and testing data were also split in an 8:1:1 ratio. In total, approximately 110,592 images were used for training, 13,824 images for validation, and 13,824 images for testing.

2) Metrics: To comprehensively evaluate the performance of the proposed CAIM-Net, we adopted five quantitative metrics: Overall Accuracy (OA), F1-score (F1), Kappa coefficient (Kappa), Precision (Pre), and Recall (Rec). The F1 score, which harmonizes precision and recall, serves as a primary evaluation metric for change detection tasks. The Kappa coefficient is particularly important in the highly imbalanced TSCD task, which involves not only detecting change areas but also identifying change moments. To evaluate the identification of change moment, we calculated the F1, Pre, and Rec for each individual change moment. These metrics were then averaged to obtain the final F1, Pre, and Rec.

3) Comparison algorithm:

We selected ten advanced methods as comparison algorithms. Among these methods, seven deep learning-based TSCD methods utilize TSIs to detect change area: LSTM, BiLSTM, UTRNet, ConvLSTM, L-UNet, MC$^2$ABNet, and RLD-Net. One deep learning-based TSCD method, Multi-RLD-Net, detects change area and identifies change moment. Additionally, two TSCD methods based on decomposition techniques, LandTrendr and CCDC, detect change area and identify change moment. Although LandTrendr and CCDC are unsupervised methods, due to the current lack of methods for detecting change moments, it is necessary to compare our proposed CAIM-Net with them.

a) LSTM (Graves, 2014): LSTM is a specialized recurrent neural network (RNN) architecture, specifically designed to address the vanishing and exploding gradient problems encountered by traditional RNNs when processing long sequence data.

b) BiLSTM (Ma and Hovy, 2016): BiLSTM is an enhanced RNN architecture that integrates two LSTM layers in opposite directions. One layer processes the forward information of the sequence, while the other processes the backward information. This bidirectional approach enables a more comprehensive understanding of contextual information.

c) UTRNet (Yang et al., 2022): UTRNet is an unsupervised convolutional recurrent network designed for change detection in irregularly collected images. It is specifically designed to distinguish pseudo-changes from real changes in TSIs.

d) ConvLSTM (Shi et al., 2015): ConvLSTM is a deep learning model that integrates convolutional layers with the LSTM architecture. By incorporating convolutional operations into the traditional LSTM, ConvLSTM can directly process spatially structured data.

e) L-UNet (Papadomanolaki et al., 2021): UNet-like architecture is a lightweight convolutional neural network specifically designed for image segmentation tasks. In TSCD, L-UNet Integrates the characteristics of LSTM networks, enabling it to effectively process time-series data and detect change regions in TSIs.

f) MC²ABNet (Li et al., 2023): MC²ABNet is a spatial-temporal-spectral joint network for change detection in medium-resolution remote sensing images. Its architecture is similar to that of L-UNet. It obtains complete temporal difference features through multi-scale convolutional channel attention and fully bidirectional ConvLSTM.

g) RLD-Net (Li and Wu, 2024): RLD-Net is a network derived from Multi-RLD-Net by removing the change moment identification branch. It primarily uses optical flow for TSIs alignment and employs LSTM to extract multi-scale difference features.

h) LandTrendr (Kennedy et al., 2010): LandTrendr is a set of spectral-temporal segmentation algorithms primarily used for change detection in time-series data of medium-resolution satellite images. It captures landscape dynamics by identifying and fitting breakpoints in TSIs.

i) CCDC (Zhu and Woodcock, 2014): CCDC is a TSIs monitoring algorithm that detects land surface changes by modeling spectral characteristics over time. It is primarily used for continuous detection and classification of land cover changes.

k) Multi-RLD-Net (Li and Wu, 2024): Multi-RLD-Net is a multi-task network for TSCD, designed to simultaneously explore change area and change moment. The network integrates optical flow and LSTM to extract differences between TSIs. Unlike the aforementioned comparison methods, LandTrendr, CCDC, and Multi-RLD-Net can identify change moment.

## 3.2. Detection performance comparison

### 3.2.1. Change detection results on DynamicEarthNet dataset

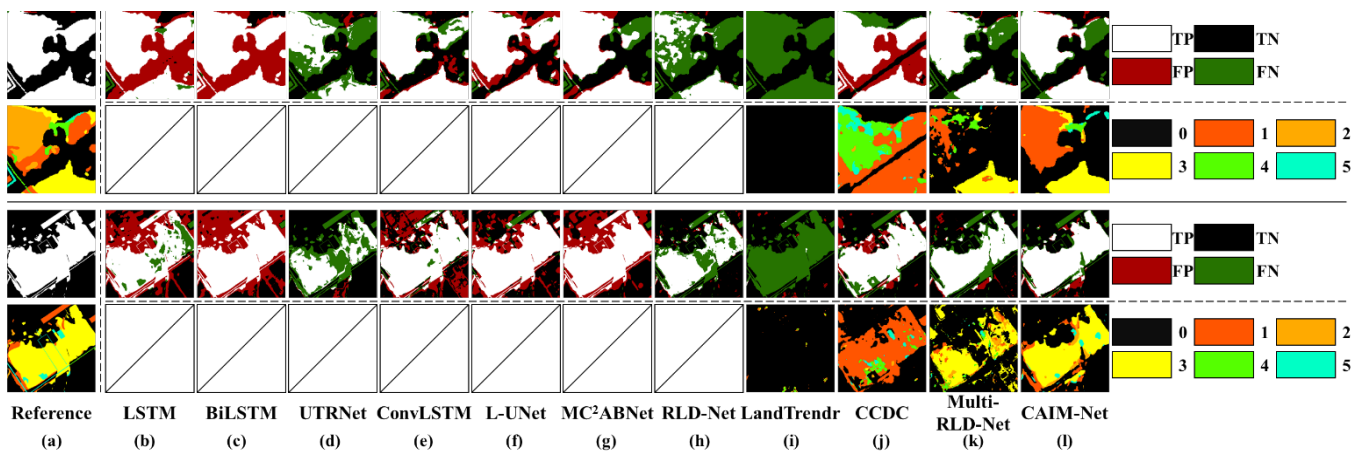

**Fig. 9**. Change area and change moment results of the DynamicEarthNet dataset. The first and third rows illustrate the change area results, where true positives are shown in white and true negatives in black, while false positives are highlighted in red and false negatives in green. The second and fourth rows illustrate the change moment results, with different colors representing

different change moments.

Table 1 Accuracy assessment for different change detection approaches on the DynamicEarthNet dataset. The table highlights the highest values in red, and the second-highest values in blue.

| | Change Area | | | | | Change Moment | | | | |
|---|---|---|---|---|---|---|---|---|---|---|
| **Method** | OA | F1 | Kappa | Pre | Rec | OA | F1 | Kappa | Pre | Rec |
| LSTM | 63.73 | 40.43 | 18.40 | 38.71 | 42.97 | / | / | / | / | / |
| BiLSTM | 66.82 | 47.21 | 24.26 | 39.70 | **58.21** | / | / | / | / | / |
| UTRNet | 76.53 | 42.16 | 28.66 | 56.92 | 33.48 | / | / | / | / | / |
| ConvLSTM | 73.10 | 49.57 | 31.27 | 47.57 | 51.76 | / | / | / | / | / |
| L-UNet | 72.70 | 50.89 | 32.16 | 47.04 | 55.42 | / | / | / | / | / |
| MC$^2$ABNet | 74.01 | 52.11 | 34.33 | 49.72 | 54.75 | / | / | / | / | / |
| RLD-Net | 80.16 | 52.88 | 41.03 | 67.25 | 43.56 | / | / | / | / | / |
| LandTrendr | 69.48 | 22.83 | 5.77 | 32.18 | 17.69 | 65.66 | 16.15 | 2.65 | 15.90 | 16.87 |
| CCDC | 80.55 | 52.77 | 41.41 | **69.51** | 42.53 | 72.22 | 22.16 | 23.46 | 26.66 | 23.07 |
| Multi-RLD-Net | 80.85 | 56.65 | 44.75 | 67.18 | 48.98 | 74.87 | 41.14 | 37.31 | 45.89 | **41.87** |
| CAIM-Net | **81.22** | **57.58** | **45.87** | 67.95 | 49.96 | **76.01** | **43.83** | **37.67** | **53.80** | 40.99 |

**Fig. 9** presents the change area detection and change moment identification results on the test set of the DynamicEarthNet dataset, comparing our framework against ten other methods. The results demonstrate that CAIM-Net yields clear and highly accurate change area and change moment maps. Among the ten comparison methods, three TSCD approaches—LSTM, BiLSTM, and UTRNet—exhibit relatively higher false positives and false negatives. This is because these methods focus solely on temporal differences without incorporating spatial information. In contrast, ConvLSTM, L-UNet, and MC$^2$ABNet simultaneously account for both spatial information and temporal differences, resulting in superior performance compared to the aforementioned patch-based methods. RLD-Net enhances change area detection by integrating optical flow into TSCD and extracting difference features at multiple scales. However, like other comparison methods, including RLD-Net, it is limited to change area detection only. On the other hand, LandTrendr, CCDC, and Multi-RLD-Net are capable of both detecting change area and identifying change moment. Nevertheless, LandTrendr, a decomposition-based method, is primarily designed for gradual change detection, while CCDC mainly identifies the moment closest to the largest magnitude change. As a result, their performance is not as satisfactory. In contrast, Multi-RLD-Net, a deep learning-based method, employs a multi-task branch to obtain detection results for both change area and change moment, positioning it as the top performer after the proposed CAIM-Net. However, our framework surpasses all these methods. The change area and change moment maps generated by our framework show very few false positive and false negative pixels, demonstrating its ability to achieve more accurate change area and change moment results by effectively information transformation between both tasks.

**Table 1** lists the overall quantitative results of our framework and the comparison methods on the test set of the

DynamicEarthNet dataset. In the change area detection task, the benchmark algorithm LSTM achieves a Kappa coefficient of 18.40%. By incorporating spatial information into TSCD, ConvLSTM, L-UNet, and MC$^2$ABNet enhance change area detection performance, increasing the Kappa coefficient by 12.87%, 13.76%, and 15.93%, respectively, compared to LSTM. RLD-Net, which integrates multiscale difference features, achieves a Kappa coefficient of 41.03%. For both change area detection and change moment identification tasks, the benchmark algorithm LandTrendr obtains Kappa coefficient of 5.77% and 2.65%, respectively. CCDC, which employs fitting techniques, improves the Kappa coefficient to 41.41% and 23.46%. Multi-RLD-Net, which utilizes a multi-task decoder to detect change area and identify change moment, achieves the second-best Kappa coefficient of 44.75% and 37.31%.

In contrast, our framework achieves the highest performance in OA, F1 score, and Kappa coefficient for both change area detection and change moment identification tasks. Compared to Multi-RLD-Net, our method demonstrates a considerable improvement in the Kappa coefficient, increasing by 1.12% and 0.36% for change area detection and change moment identification, respectively. Multi-RLD-Net, one of the most advanced TSCD algorithms, achieves the second-highest accuracy on the DynamicEarthNet dataset. These results fully demonstrate the superiority of our method for both tasks.

### 3.2.2. Change detection results on SpaceNet7 dataset

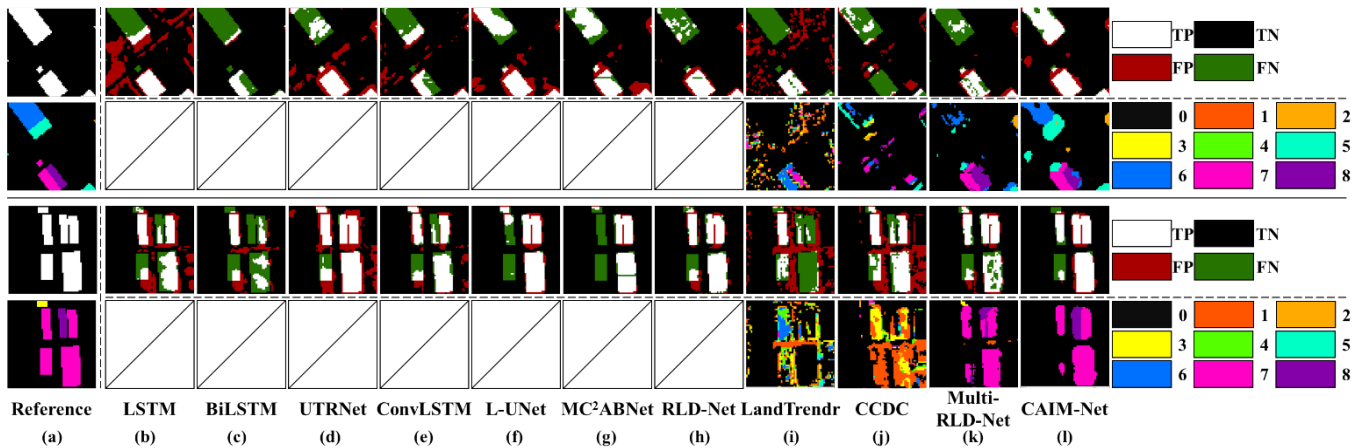

**Fig. 10**. Change area and change moment results of the SpaceNet7 dataset. The first and third rows illustrate the change area results, where true positives are shown in white and true negatives in black, while false positives are highlighted in red and false negatives in green. The second and fourth rows illustrate the change moment results, with different colors representing different change moments.

**Fig. 10** presents change detection results on the test set of the SpaceNet7 dataset. Compared to the DynamicEarthNet dataset, SpaceNet7 dataset is highly imbalanced, with changed samples constituting only 1% of the total samples. This imbalance poses significant challenges for change detection, particularly for the TSCD task, which requires not only detecting change area but also identifying the precise moment of change. Existing methods exhibit various limitations. First, methods designed for detecting change area, such as LSTM, BiLSTM, UTRNet, ConvLSTM, L-UNet, MC$^2$ABNet, and RLD-Net, suffer from false positives and false negatives to varying degrees. Second, methods capable of both change area detection and change moment identification, such as LandTrendr and CCDC, capture changes at the single-pixel level over time, resulting in cluttered and disorganized change maps.

Moreover, their visualization results often fail to accurately pinpoint the moment of change. Compared with these methods, Multi-RLD-Net performs better in detecting change area and identifying change moment but still has a notable number of false negatives. In contrast, our proposed framework achieves the best performance in both tasks. By incorporating spatial information, it produces more structured visualization results, where changed and unchanged pixels form coherent patches. Additionally, it has the fewest false positives and false negatives and accurately identifies the corresponding change moment.

**Table 2** Accuracy assessment for different change detection approaches on the SpaceNet7 dataset. The table highlights the highest values in red, and the second-highest values in blue.

| | Change Area | | | | | Change Moment | | | | |
|---|---|---|---|---|---|---|---|---|---|---|
| **Method** | OA | F1 | Kappa | Pre | Rec | OA | F1 | Kappa | Pre | Rec |
| LSTM | 94.30 | 32.03 | 29.70 | 21.97 | 59.04 | / | / | / | / | / |
| BiLSTM | 96.77 | 32.78 | 31.13 | 31.13 | 34.61 | / | / | / | / | / |
| UTRNet | 94.15 | 35.63 | 33.35 | 23.76 | 71.14 | / | / | / | / | / |
| ConvLSTM | 96.08 | 36.56 | 34.69 | 28.93 | 49.66 | / | / | / | / | / |
| L-UNet | 97.69 | 46.18 | 45.00 | 49.06 | 43.61 | / | / | / | / | / |
| MC$^2$ABNet | 97.96 | 48.18 | 47.17 | 57.02 | 41.72 | / | / | / | / | / |
| RLD-Net | 97.70 | 50.10 | 48.92 | 49.41 | 50.80 | / | / | / | / | / |
| LandTrendr | 80.30 | 4.42 | 0.36 | 2.49 | 19.90 | 79.93 | 10.38 | 0.35 | 11.14 | 12.43 |
| CCDC | 91.24 | 16.18 | 13.07 | 10.36 | 36.94 | 90.50 | 12.39 | 7.24 | 12.16 | 14.51 |
| Multi-RLD-Net | 97.68 | 50.51 | 49.33 | 49.15 | 51.96 | 97.63 | 41.84 | 43.64 | 47.02 | 39.83 |
| CAIM-Net | 97.97 | 52.53 | 51.49 | 55.97 | 49.48 | 97.66 | 40.73 | 44.61 | 44.74 | 39.57 |

The quantitative results of our framework and comparison methods are reported in **Table 2**. The SpaceNet7 dataset, which focuses solely on building change and has a higher spatial resolution than the DynamicEarthNet dataset, leads to higher accuracy for all methods. The Kappa coefficient for the benchmark method is 29.70%. UTRNet, which employs a time-distance-guided LSTM, detects difference features more effectively than BiLSTM, resulting in a 2.22% improvement in the Kappa coefficient over BiLSTM on the SpaceNet7 dataset. Compared to the DynamicEarthNet dataset, the performance improvement of ConvLSTM over the benchmark LSTM is less pronounced, whereas L-UNet and MC$^2$ABNet exhibit more substantial gains. For both change area detection and change moment identification tasks, the relatively fixed model structures of LandTrendr and CCDC make it difficult for them to adapt to the diversity and complexity of building changes in high-resolution images, leading to poor performance on the SpaceNet7 dataset. Multi-RLD-Net achieves the second-highest Kappa coefficient of 49.33% and 43.63%. By leveraging deep learning to capture spatiotemporal features, it better adapts to various types of building changes.

Finally, our CAIM-Net achieved the highest Kappa coefficient of 51.49% and 44.61% for both tasks, representing improvements of 2.16% and 0.97% compared to the SOTA method, Multi-RLD-Net. It is worth noting that, for the change moment identification task on the SpaceNet7 dataset, the F1 score of our proposed CAIM-Net is lower than that of Multi-RLD-Net. Given the extreme class imbalance in the SpaceNet7 dataset, we argue that the Kappa coefficient is a more reliable metric, as it better reflect the model's true performance on the minority class. The superiority of our framework can be attributed to several factors: (1) Boundary enhancement convolution enables the framework to extract more precise difference features, thereby improving the accuracy of change detection. (2) Utilization of change and no-change features between adjacent images in TSIs, our framework improves the accuracy of change moment identification. (3) Collaborative detection of change area and change moment—CAIM-Net leverages the inherent relationship between these two tasks to infer change area from change moment, leading to simultaneous improvements in both.

The comparisons on both datasets demonstrate the superiority of our proposed framework in detecting change area across diverse scenes. These results also validate our motivation to infer change area from change moment by leveraging the inherent property between change area and change moment.

### 3.3. Ablation study

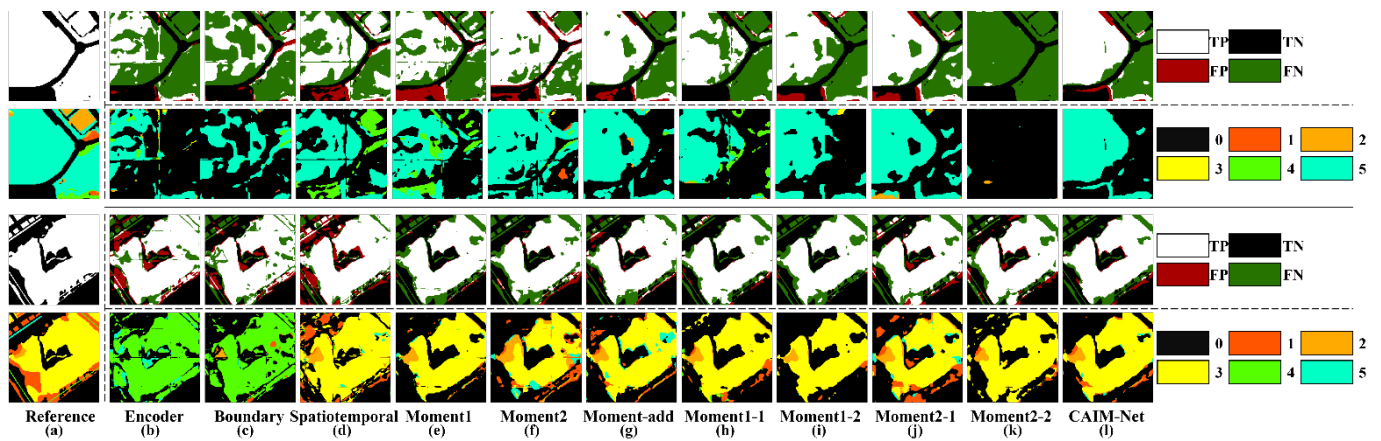

**Fig. 11.** The ablation results of DynamicEarthNet dataset. The first and third rows illustrate the change area results, where true positives are shown in white and true negatives in black, while false positives are highlighted in red and false negatives in green. The second and fourth rows illustrate the change moment results, with different colors representing different change moments.

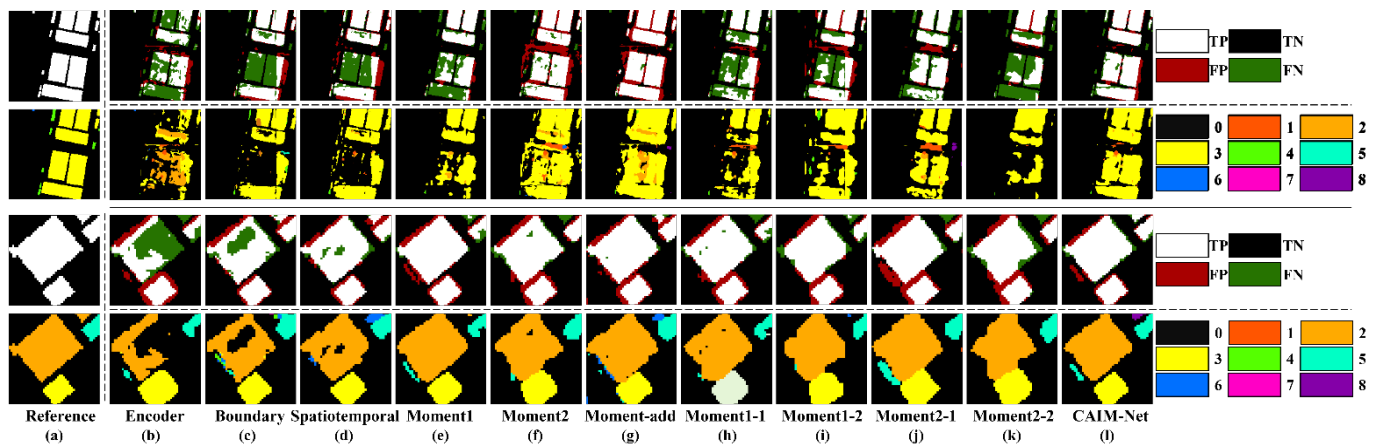

**Fig. 12.** The ablation results of SpaceNet7 dataset. The first and third rows illustrate the change area results, where true positives are shown in white and true negatives in black, while false positives are highlighted in red and false negatives in green. The second and fourth rows illustrate the change moment results, with different colors representing different change moments.

**Table 3** Ablation experimental results of the proposed framework on the DynamicEarthNet dataset.

| Method | Change Area | | | | | Change Moment | | | | |
|---|---|---|---|---|---|---|---|---|---|---|
| | OA | F1 | Kappa | Pre | Rec | OA | F1 | Kappa | Pre | Rec |
| Encoder | 75.15 | 48.00 | 31.77 | 51.50 | 44.95 | 68.60 | 23.52 | 23.23 | 38.71 | 29.28 |
| Boundary | 75.32 | 48.49 | 32.35 | 51.87 | 45.52 | 69.23 | 25.80 | 24.91 | 37.27 | 31.28 |
| Spatiotemporal | 71.26 | 53.45 | 33.56 | 45.55 | **64.68** | 63.48 | 36.46 | 27.90 | 33.78 | *41.64* |
| Moment1 | 79.65 | 49.35 | 37.76 | 67.64 | 38.84 | *76.28* | 40.54 | 33.52 | 50.47 | 38.27 |
| Moment2 | 79.14 | **57.97** | *44.11* | 59.66 | *56.36* | 72.55 | 39.73 | 34.96 | 46.32 | 40.33 |
| Moment1+2 | 79.13 | 55.34 | 41.87 | 60.95 | 50.68 | 73.96 | 42.59 | 35.22 | 48.35 | 41.00 |
| Moment1-1 | *80.18* | 53.74 | 41.72 | 66.46 | 45.11 | 75.67 | 39.41 | 35.21 | *50.68* | 39.09 |
| Moment1-2 | 80.05 | 48.84 | 37.89 | **70.66** | 37.32 | **77.25** | 40.58 | 34.78 | 52.40 | 39.39 |
| Moment2-1 | 78.31 | 56.97 | 42.47 | 57.69 | 56.27 | 72.63 | *42.80* | *35.96* | 44.15 | **43.00** |
| Moment2-2 | 78.74 | 52.83 | 39.42 | 60.89 | 46.66 | 73.79 | 38.18 | 32.92 | 48.06 | 38.82 |
| CAIM-Net | **81.22** | *57.58* | **45.87** | *67.95* | 49.96 | 76.01 | **43.83** | **37.67** | **53.80** | 40.99 |

**Table 4** Ablation experimental results of the proposed framework on the SpaceNet7 dataset.

| Method | Change Area | | | | | Change Moment | | | | |
|---|---|---|---|---|---|---|---|---|---|---|
| | OA | F1 | Kappa | Pre | Rec | OA | F1 | Kappa | Pre | Rec |
| Encoder | 97.66 | 42.22 | 41.05 | 48.13 | 37.61 | 97.46 | 38.83 | 36.67 | 43.42 | 37.20 |
| Boundary | 97.74 | 44.66 | 43.52 | 50.29 | 40.16 | 95.54 | 40.33 | 39.28 | 44.23 | 38.74 |
| Spatiotemporal | 97.30 | 47.80 | 46.43 | 42.64 | *54.38* | 96.96 | 39.07 | 40.36 | 41.49 | **43.60** |
| Moment1 | 97.69 | 49.25 | 48.06 | 49.15 | 49.34 | 97.32 | 38.31 | 40.50 | 39.62 | 39.58 |
| Moment2 | 97.86 | 48.47 | 47.39 | 53.61 | 44.24 | 97.61 | 40.59 | 41.70 | *46.94* | 38.49 |
| Moment1+2 | 97.57 | *51.48* | *50.25* | 47.20 | **56.63** | 97.19 | 39.19 | 43.13 | 42.99 | *42.39* |
| Moment1-1 | 97.76 | 50.63 | 49.48 | 50.73 | 50.53 | 97.46 | *41.30* | *43.30* | 44.91 | 41.52 |
| Moment1-2 | 97.74 | 48.58 | 47.42 | 50.26 | 47.00 | 97.48 | **41.71** | 42.14 | 43.68 | 41.09 |
| Moment2-1 | 97.73 | 50.29 | 49.13 | 50.05 | 50.53 | 97.42 | 39.00 | 42.91 | 44.77 | 40.11 |
| Moment2-2 | *97.87* | 49.31 | 48.24 | *53.89* | 45.46 | *97.61* | 40.35 | 42.43 | **47.65** | 37.70 |
| CAIM-Net | **97.97** | **52.53** | **51.49** | **55.97** | 49.48 | **97.66** | 40.73 | **44.61** | 44.74 | 39.57 |

Our framework comprises six key components: i.e., a spatial features extractor, boundary enhancement convolution, a spatiotemporal correlation module, change moment extractor1, change moment extractor2, and a multiscale temporal CAM. To validate the effectiveness of each component, we conduct an ablation study on two benchmark datasets and document the contribution of each component to the final detection and identification in **Table 3** and **Table 4**. In **Table 3** and **Table 4**, 'Encoder' refers to the spatial feature extractor, which extracts spatial features and then employs a simple method to identify the change moment. 'Boundary' indicates the addition of the boundary enhancement convolution. 'Spatiotemporal' signifies the incorporation of the spatiotemporal correlation module. 'Moment1' demotes the addition of change moment extractor1. 'Moment2' denotes the addition of change moment extractor2. 'Moment1+2' means that the coarse change moments obtained by the two change moment extractors are summed to produce the final change moment. 'Moment1-1' indicates that the temporal CAM with scale 1 is applied to coarse change moment1 to obtain a refined change moment. 'Moment1-2' indicates that the temporal CAM with scale 2 is applied to coarse change moment1, 'Moment2-1' indicates that the temporal CAM with scale 1 is applied to coarse change moment2, 'Moment2-2' indicates that the temporal CAM with scale 2 is applied on coarse change moment2. 'CAIM-Net' represents the final proposed framework, which incorporates the multiscale temporal CAM.

Firstly, utilizing only the spatial feature extractor and a simple decoder to identify the change moment yields Kappa coefficient of 31.77% and 23.23% on the DynamicEarthNet dataset, and 41.05% and 36.67% on the SpaceNet7 dataset respectively. These values outperform patch-based SOTA methods such as BiLSTM and UTRNet. Compared to the simple spatial features extractor, the boundary enhancement convolution generates more distinct and clear difference features for subsequent change detection. As a result, the Kappa coefficient for the change area and change moment improves to 32.35% and 24.91% on the DynamicEarthNet dataset and 43.52% and 39.28% on the SpaceNet7 dataset, respectively. The introduction of the spatiotemporal module further enhances performance, resulting in Kappa coefficient of 33.56% and 27.90% on the DynamicEarthNet dataset, and 46.43% and 40.36% on the SpaceNet7 dataset. These findings indicate that the boundary enhancement convolution and spatiotemporal module effectively generate more distinct difference features for change moment identification. Next, the designed change moment extractor1 and change moment extractor2 are used to generate coarse change moments. Specifically, change moment extractor1 improves the Kappa coefficient for the change area and change moment to 37.76% and 33.52% on the DynamicEarthNet dataset and 48.06% and 40.50% on the SpaceNet7 dataset. Change moment extractor 2 enhances these values to 44.11% and 34.96% on the DynamicEarthNet dataset and 47.39% and 41.70% on the SpaceNet7 dataset. By incorporating coarse change moment extractor1 and coarse change moment extractor2 together, the performance is further enhanced, reaching 41.87% and 35.22% on the DynamicEarthNet dataset and 50.25% and 43.13% on the SpaceNet dataset. The multiscale temporal CAM effectively determines the exact moment when these changes occur, with different scale exhibiting varying degrees of refinement. For example, Moment1-1 improves the kappa coefficient for change are and change moment from 37.76% and 33.52% to 41.72% and 35.21% on the DynamicEarthNet dataset, and from 48.06% and 40.50% to 49.48% and 43.30% on the SpaceNet7 dataset. Finally, incorporating the multiscale temporal CAM together significantly enhances the overall performance of the framework. The final Kappa coefficient on the two datasets reaches 45.87% and 37.67% on the DynamicEarthNet dataset, and 51.49% and 44.61% on the SpaceNet7 dataset. Notably, throughout this process, we leverage the inherent relationship between change area and change moment to infer change area from the identified change moment. As a result, the generation of the change area is fully automated.

**Fig. 11** and **Fig. 12** illustrate ablation results on the DynamicEarthNet and SpaceNet7 datasets. In **Fig. 11**, we observe that while the spatial features extractor and boundary enhancement convolution can detect some change areas, they cannot accurately identify the change moment. After incorporating the spatiotemporal correlation module, which utilizes an LSTM to capture the temporal correlation of difference features, the framework can more accurately identify the change moment. This indicates the importance of temporal information for change moment identification. In **Fig. 12**, we find that both change moment extractor1 and change moment extractor2 exbibit false positives and false negatives to varying degrees. However, by aggregating change moments across multiple scales temporal CAMs enables the framework to integrate information from different branches and scales more effectively. This strategy reduces false positives and false negatives, leading to more accurate change moment and change area.

## 4. Discussion

### 4.1. Discussion about the multi branch outputs

Table 5 Multi branch outputs of the proposed framework on the DynamicEarthNet dataset.

|  | Change Area | | | | | Change Moment | | | | |
| --- | --- | --- | --- | --- | --- | --- | --- | --- | --- | --- |
| **Method** | **OA** | **F1** | **Kappa** | **Pre** | **Rec** | **OA** | **F1** | **Kappa** | **Pre** | **Rec** |
| Moment1 | 80.66 | 57.20 | 44.97 | 65.71 | 50.64 | 75.72 | 45.21 | 37.90 | 54.11 | 41.92 |
| Moment2 | 80.95 | 55.89 | 44.25 | 68.28 | 47.31 | 76.41 | 44.51 | 37.50 | 54.51 | 40.80 |
| Moment3 | 79.96 | 57.33 | 44.36 | 62.76 | **52.76** | 73.95 | 41.65 | 35.43 | 49.74 | **40.36** |
| Moment4 | **80.23** | 56.19 | 43.71 | **64.65** | 49.70 | 74.55 | 40.84 | 34.89 | **50.18** | 39.23 |
| CAIM-Net | **81.22** | **57.58** | **45.87** | **67.95** | 49.96 | 76.01 | **43.83** | **37.67** | **53.80** | 40.99 |

The proposed framework not only provides the final change moment output but also generates four additional change moment outputs: Moment1, Moment2, Moment3, and Moment4. To evaluate the effectiveness of training multiple branches together, we conducted a quantitative assessment of each branch's output, with the results summarized in **Table 5**. The findings indicate that multi-branch training significantly enhances overall performance. Notably, the Kappa coefficients for Moment1 and Moment2 branches in change moment identification reached 37.9% and 37.5%, respectively, comparable to that of the final change moment. These results demonstrate that training multiple branches together improves individual branch performance, leading to more accurate change detection.

### 4.2. Discussion about the run time

Table 6 Train and inference time of the Siamese structure and proposed Encoder on the DynamicEarthNet dataset.

|  | Train Time | Inference Time |
| --- | --- | --- |
| Siamese | 292.83 | 9.62 |
| Encoder | 46.49 | 6.13 |

Table 7 Train and inference time of the Siamese structure and proposed Encoder on the SpaceNet7 dataset.

|  | Train Time | Inference Time |
| --- | --- | --- |
| Siamese | 1281.02 | 26.80 |
| Encoder | 134.09 | 14.40 |

we adopted a training method that merges the time dimension with the batch dimension to extract spatial features,

significantly accelerates training. To evaluate the computational efficiency of this approach, we compared it with the traditional siamese structure on the DynamicEarthNet and SpaceNet7 datasets, as shown in **Table 6** and **Table 7**. Specifically, the siamese structure requires 292.83s and 1281.02s per epoch for training on the two datasets, whereas our proposed Encoder takes only 46.69s and 134.09s per epoch—approximately 16% and 10% of the time required by the siamese structure. During inference, the siamese structure takes 9.62s and 26.8s to extract spatial features, while our Encoder requires just 6.13s and 14.4s, which is 64% and 54% of the time needed by the siamese structure. This smaller gap in inference time is due to the fact that inference involves only forward propagation, without backpropagation and parameter updates.

The commonly used siamese structure processes each time step independently using the same spatial feature extraction module, essentially forming a parameter-sharing recurrent structure. In contrast, the Encoder in this study merges time steps and batches into a single input for spatial feature extractor while still sharing parameters. Theoretically, both approaches are equivalent in terms of feature extraction. However, the Encoder's large-batch processing is more efficient, as it consolidates multiple time steps and batches into a single forward pass. Despite this efficiency gain, the model parameters remain unchanged, leading to similar extracted features. Therefore, the Encoder not only preserves the accuracy of spatial feature extraction but also significantly reduces training time.

**4.3. Discussion about the efficiency**

Table 8 FLOPs, parameters, and inference time of different methods on SpaceNet7 dataset.

|  | Flops (G) | Parameters (M) | Inference Time (s) |
|---|---|---|---|
| LSTM | 12.10 | 0.0049 | 3.92 |
| BiLSTM | 14.51 | 0.0058 | 3.62 |
| UTRNet | 350.85 | 0.0721 | 22.20 |
| ConvLSTM | 0.16 | 0.0035 | 5.87 |
| L-UNet | 1078.50 | 8.4450 | 9.59 |
| MC$^2$ABNet | 13189.78 | 94.1032 | 24.33 |
| RLD-Net | 156.56 | 0.0801 | 170.94 |
| Multi-RLD-Net | 156.59 | 0.0812 | 161.42 |
| CAIM-Net | 683.62 | 2.2262 | 19.92 |

FLOPs, parameters, and inference time are crucial metrics for evaluating the efficiency and practicality of change detection models. **Table 8** presents a comparison of these metrics across various deep learning-based methods and the proposed CAIM-Net on the SpaceNet7 dataset.

FLOPs are primarily measure model complexity and computational resource requirements. Among the comparison methods, LSTM, BiLSTM, UTRNet, ConvLSTM, RLD-Net, and Multi-RLD-Net exhibit relatively low complexity. In contrast, L-UNet and MC$^2$ABNet have significantly higher FLOPs, with MC$^2$ABNet being particularly computationally intensive—its FLOPs are nearly ten times greater than those of the second-highest model, L-UNet. Our proposed CAIM-Net has higher FLOPs than pixel- and patch-based methods but remains significantly lower than L-UNet and MC$^2$ABNet. Specifically, CAIM-Net's FLOPs amount to only 5% of those of MC$^2$ABNet's, indicating substantially lower computational complexity. Parameters serve as another key indicator of model complexity and storage requirements. The parameter counts of the comparison methods follow a similar trend to the FLOPs. Notably, CAIM-Net's parameter count is only 2% of MC$^2$ABNet's, further demonstrating its reduced complexity and storage demands.

Inference time reflects the model's real-time performance and efficiency. Pixel-based methods, such as LSTM and BiLSTM feature simple architecture, resulting in shorter inference times. Conversely, methods like UTRNet, L-UNet, and MC$^2$ABNet have more complex structures, leading to moderate inference times. Notably, RLD-Net and Multi-RLD-Net exhibit significantly longer inference times—up to 160 seconds—due to their patch-based input architecture, which demands extensive computation during both training and inference. In contrast, CAIM-Net requires only 19.92s for inference, demonstrating its ability to achieve accurate change area detection and change moment identification in substantially less time. This efficiency significantly reduces both training and inference time.

## 5. Conclusion

This article addresses two primary challenges in the field of TSCD: (1) unclear change boundaries due to the blurred characteristics of objects, and (2) the mismatch between the change area and change moment. To tackle these challenges, we propose CAIM-Net, a novel framework that consists of three key steps: Difference Extraction and Enhancement, Coarse Change Moment Extraction, and Fine Change Moment Extraction and Change Area Inference. The Difference Extraction and Enhancement step is designed to capture more refined and effective difference features from TSIs, facilitating subsequent change area detection and change moment identification. The Coarse Change Moment Extraction step leverages these difference features to determine the initial change moment. Here, we introduce a novel method that infers the change moment based on change and no-change features between adjacent images in TSIs. Finally, the Fine Change Moment Extraction and Change Area Inference step exploits the principle that pixels with an identified change moment must have undergone a change. This step enhances the accuracy of both the change moment and change area detection by leveraging their interdependence CAIM-Net demonstrates a powerful capability for extracting and utilizing difference features to accurately identify change moment and infer change area. Experiments on two TSIs datasets validate its effectiveness. Compared to existing SOTA models, CAIM-Net significantly improves the accuracy and robustness of change detection and identification. Additionally, our method exhibits superior runtime efficiency, as demonstrated in the experiments.

Despite its advanced performance in change area detection and change moment identification, CAIM-Net has certain limitations. It focuses primarily on detecting the moment of the last change event, overlooking the intermediate changes between temporally adjacent images. However, capturing these intermediate changes would provide a more comprehensive understanding of spatiotemporal dynamics, which is crucial for practical applications. Therefore, in future work, we will focus on detecting and characterizing changes between every pair of temporally adjacent images within TSIs.

**Acknowledgements**

This work was supported in part by the National Key Research and Development Program of China under Grant

2022YFB3903300, and in part by National Natural Science Foundation of China under Grant T2122014. The numerical calculations in this paper have been done on the supercomputing system in the Supercomputing Center of Wuhan University.2022YFB3903300, and in part by National Natural Science Foundation of China under Grant T2122014. The numerical calculations in this paper have been done on the supercomputing system in the Supercomputing Center of Wuhan University.

Reference

Anno, S., Tsubasa, H., Sugita, S., Yasumoto, S., Lee, M.-A., Sasaki, Y., Oyoshi, K., 2024. Challenges and implications of predicting the spatiotemporal distribution of dengue fever outbreak in Chinese Taiwan using remote sensing data and deep learning. Geo-Spat. Inf. Sci. 27, 1155–1161. https://doi.org/10.1080/10095020.2022.2144770

Bovolo, F., Bruzzone, L., 2007. A Theoretical Framework for Unsupervised Change Detection Based on Change Vector Analysis in the Polar Domain. IEEE Trans. Geosci. Remote Sens. 45, 218–236. https://doi.org/10.1109/TGRS.2006.885408

Chakraborty, S., Stokes, E.C., 2023. Adaptive modeling of satellite-derived nighttime lights time-series for tracking urban change processes using machine learning. Remote Sens. Environ. 298, 113818. https://doi.org/10.1016/j.rse.2023.113818

Chen, G., Zhou, Y., Voogt, J.A., Stokes, E.C., 2024. Remote sensing of diverse urban environments: From the single city to multiple cities. Remote Sens. Environ. 305, 114108. https://doi.org/10.1016/j.rse.2024.114108

Chen, H., Song, J., Han, C., Xia, J., Yokoya, N., 2024a. ChangeMamba: Remote Sensing Change Detection With Spatiotemporal State Space Model. IEEE Trans. Geosci. Remote Sens. 62, 1–20. https://doi.org/10.1109/TGRS.2024.3417253

Chen, H., Song, J., Wu, C., Du, B., Yokoya, N., 2023. Exchange means change: An unsupervised single-temporal change detection framework based on intra- and inter-image patch exchange. ISPRS J. Photogramm. Remote Sens. 206, 87–105. https://doi.org/10.1016/j.isprsjprs.2023.11.004

Chen, H., Song, J., Yokoya, N., 2024b. Change Detection Between Optical Remote Sensing Imagery and Map Data via Segment Anything Model (SAM). https://doi.org/10.48550/arXiv.2401.09019

Chen, S., Woodcock, C.E., Bullock, E.L., Arévalo, P., Torchinava, P., Peng, S., Olofsson, P., 2021. Monitoring temperate forest degradation on Google Earth Engine using Landsat time series analysis. Remote Sens. Environ. 265, 112648. https://doi.org/10.1016/j.rse.2021.112648

Chen, T.-H.K., Pandey, B., Seto, K.C., 2023. Detecting subpixel human settlements in mountains using deep learning: A case of the Hindu Kush Himalaya 1990–2020. Remote Sens. Environ. 294, 113625. https://doi.org/10.1016/j.rse.2023.113625

Cheng, X., Li, Z., 2024. Modeling information flow from multispectral remote sensing images to land use and land cover maps for understanding classification mechanism. Geo-Spat. Inf. Sci. 27, 1568–1584. https://doi.org/10.1080/10095020.2023.2275625

Deng, J.S., Wang, K., Deng, Y.H., Qi, G.J., 2008. PCA-based land-use change detection and analysis using multitemporal and multisensor satellite data. Int. J. Remote Sens. 29, 4823–4838. https://doi.org/10.1080/01431160801950162

Dong, J., Zhang, T., Wang, L., Li, Zhengqiang, Sing Wong, M., Bilal, M., Zhu, Z., Mao, F., Xia, X., Han, G., Xu, Q., Gu, Y., Lin, Y., Zhao, B., Li, Zhiwei, Xu, K., Chen, X., Gong, W., 2024. First retrieval of daily 160 m aerosol optical depth over urban areas using Gaofen-1/6 synergistic observations: Algorithm development and validation. ISPRS J. Photogramm. Remote Sens. 211, 372–391. https://doi.org/10.1016/j.isprsjprs.2024.04.020

Feng, J., Yang, X., Gu, Z., 2024. SGNet: A Transformer-Based Semantic-Guided Network for Building Change Detection. IEEE J. Sel. Top. Appl. Earth Obs. Remote Sens. 17, 9922–9935. https://doi.org/10.1109/JSTARS.2024.3402388

Graves, A., 2014. Generating Sequences With Recurrent Neural Networks. https://doi.org/10.48550/arXiv.1308.0850

Han, C., Wu, C., Guo, H., Hu, M., Li, J., Chen, H., 2023. Change Guiding Network: Incorporating Change Prior to Guide Change Detection in Remote Sensing Imagery. IEEE J. Sel. Top. Appl. Earth Obs. Remote Sens. 16, 8395–8407. https://doi.org/10.1109/JSTARS.2023.3310208

He, K., Zhang, X., Ren, S., Sun, J., 2016. Deep Residual Learning for Image Recognition, in: 2016 IEEE Conference on Computer Vision and Pattern Recognition (CVPR). Presented at the 2016 IEEE Conference on Computer Vision and Pattern Recognition (CVPR), IEEE, Las Vegas, NV, USA, pp. 770–778. https://doi.org/10.1109/CVPR.2016.90

Hou, H., Shen, L., Jia, J., Xu, Z., 2024. An integrated framework for flood disaster information extraction and analysis leveraging social media data: A case study of the Shouguang flood in China. Sci. Total Environ. 949, 174948.

https://doi.org/10.1016/j.scitotenv.2024.174948

Kennedy, R.E., Yang, Z., Cohen, W.B., 2010. Detecting trends in forest disturbance and recovery using yearly Landsat time series: 1. LandTrendr — Temporal segmentation algorithms. Remote Sens. Environ. 114, 2897–2910. https://doi.org/10.1016/j.rse.2010.07.008

Kingma, D.P., Ba, J., 2017. Adam: A Method for Stochastic Optimization. https://doi.org/10.48550/arXiv.1412.6980

Li, J., Hu, M., Wu, C., 2023. Multiscale Change Detection Network Based on Channel Attention and Fully Convolutional BiLSTM for Medium-Resolution Remote Sensing Imagery. IEEE J. Sel. Top. Appl. Earth Obs. Remote Sens. 16, 9735–9748. https://doi.org/10.1109/JSTARS.2023.3323372

Li, J., Wu, C., 2024. Using difference features effectively: A multi-task network for exploring change areas and change moments in time series remote sensing images. ISPRS J. Photogramm. Remote Sens. 218, 487–505. https://doi.org/10.1016/j.isprsjprs.2024.09.029

Li, K., Cao, X., Meng, D., 2024. A New Learning Paradigm for Foundation Model-Based Remote-Sensing Change Detection. IEEE Trans. Geosci. Remote Sens. 62, 1–12. https://doi.org/10.1109/TGRS.2024.3365825

Lian Zhao, Jinshui Zhang, 2011. Remote sensing recognition winter wheat with change vector analysis (CVA) based on cultivated parcel data, in: 2011 International Conference on Remote Sensing, Environment and Transportation Engineering. Presented at the 2011 International Conference on Remote Sensing, Environment and Transportation Engineering (RSETE), IEEE, Nanjing, China, pp. 5218–5224. https://doi.org/10.1109/RSETE.2011.5965490

Liu, Z., He, D., Shi, Q., Cheng, X., 2024. NDVI time-series data reconstruction for spatial-temporal dynamic monitoring of Arctic vegetation structure. Geo-Spat. Inf. Sci. 1–19. https://doi.org/10.1080/10095020.2024.2336602

Lou, C., Al-qaness, M.A.A., AL-Alimi, D., Dahou, A., Abd Elaziz, M., Abualigah, L., Ewees, A.A., 2024. Land use/land cover (LULC) classification using hyperspectral images: a review. Geo-Spat. Inf. Sci. 1–42. https://doi.org/10.1080/10095020.2024.2332638

Ma, X., Hovy, E., 2016. End-to-end Sequence Labeling via Bi-directional LSTM-CNNs-CRF. https://doi.org/10.48550/arXiv.1603.01354

Masoumi, Z., Van Genderen, J., 2024. Artificial intelligence for sustainable development of smart cities and urban land-use management. Geo-Spat. Inf. Sci. 27, 1212–1236. https://doi.org/10.1080/10095020.2023.2184729

Ng, M.K., Yau, A.C., 2005. Super-Resolution Image Restoration from Blurred Low-Resolution Images. J. Math. Imaging Vis. 23, 367–378. https://doi.org/10.1007/s10851-005-2028-5

Ojala, T., Pietikainen, M., Harwood, D., 1994. Performance evaluation of texture measures with classification based on Kullback discrimination of distributions, in: Proceedings of 12th International Conference on Pattern Recognition. Presented at the 12th International Conference on Pattern Recognition, IEEE Comput. Soc. Press, Jerusalem, Israel, pp. 582–585. https://doi.org/10.1109/ICPR.1994.576366

Pal, S.K., Majumdar, T.J., Bhattacharya, A.K., 2007. ERS-2 SAR and IRS-1C LISS III data fusion: A PCA approach to improve remote sensing based geological interpretation. ISPRS J. Photogramm. Remote Sens. 61, 281–297. https://doi.org/10.1016/j.isprsjprs.2006.10.001

Papadomanolaki, M., Vakalopoulou, M., Karantzalos, K., 2021. A Deep Multitask Learning Framework Coupling Semantic Segmentation and Fully Convolutional LSTM Networks for Urban Change Detection. IEEE Trans. Geosci. Remote Sens. 59, 7651–7668. https://doi.org/10.1109/TGRS.2021.3055584

Peng, Z., Jiang, D., Li, W., Mu, Q., Li, X., Cao, W., Shi, Z., Chen, T., Huang, J., 2024. Impacts of the scale effect on quantifying the response of spring vegetation phenology to urban intensity. Remote Sens. Environ. 315, 114485. https://doi.org/10.1016/j.rse.2024.114485

Qin, X., Guo, H., Su, X., Zhao, Z., Wang, D., Zhang, L., 2025. Spatiotemporal masked pre-training for advancing crop mapping on satellite image time series with limited labels. Int. J. Appl. Earth Obs. Geoinformation 137, 104426. https://doi.org/10.1016/j.jag.2025.104426

Runge, A., Nitze, I., Grosse, G., 2022. Remote sensing annual dynamics of rapid permafrost thaw disturbances with LandTrendr. Remote Sens. Environ. 268, 112752. https://doi.org/10.1016/j.rse.2021.112752

Shen, H., Meng, X., Zhang, L., 2016. An Integrated Framework for the Spatio–Temporal–Spectral Fusion of Remote Sensing Images. IEEE Trans. Geosci. Remote Sens. 54, 7135–7148. https://doi.org/10.1109/TGRS.2016.2596290

Shi, X., Chen, Z., Wang, H., Yeung, D.-Y., Wong, W., Woo, W., 2015. Convolutional LSTM Network: A Machine Learning

Approach for Precipitation Nowcasting. https://doi.org/10.48550/arXiv.1506.04214

Simonyan, K., Zisserman, A., 2015. Very Deep Convolutional Networks for Large-Scale Image Recognition. https://doi.org/10.48550/arXiv.1409.1556

Sun, L., Lou, Y., Shi, Q., Zhang, L., 2024. Spatial domain transfer: Cross-regional paddy rice mapping with a few samples based on Sentinel-1 and Sentinel-2 data on GEE. Int. J. Appl. Earth Obs. Geoinformation 128, 103762. https://doi.org/10.1016/j.jag.2024.103762

Toker, A., Kondmann, L., Weber, M., Eisenberger, M., Camero, A., Hu, J., Hoderlein, A.P., Senaras, C., Davis, T., Cremers, D., Marchisio, G., Zhu, X.X., Leal-Taixe, L., 2022. DynamicEarthNet: Daily Multi-Spectral Satellite Dataset for Semantic Change Segmentation, in: 2022 IEEE/CVF Conference on Computer Vision and Pattern Recognition (CVPR). Presented at the 2022 IEEE/CVF Conference on Computer Vision and Pattern Recognition (CVPR), IEEE, New Orleans, LA, USA, pp. 21126–21135. https://doi.org/10.1109/CVPR52688.2022.02048

Tong, X.-Y., Dong, R., Zhu, X.X., 2025. Global high categorical resolution land cover mapping via weak supervision. ISPRS J. Photogramm. Remote Sens. 220, 535–549. https://doi.org/10.1016/j.isprsjprs.2024.12.017

Van Etten, A., Hogan, D., Manso, J.M., Shermeyer, J., Weir, N., Lewis, R., 2021. The Multi-Temporal Urban Development SpaceNet Dataset, in: 2021 IEEE/CVF Conference on Computer Vision and Pattern Recognition (CVPR). Presented at the 2021 IEEE/CVF Conference on Computer Vision and Pattern Recognition (CVPR), IEEE, Nashville, TN, USA, pp. 6394–6403. https://doi.org/10.1109/CVPR46437.2021.00633

Wang, Z., Yi, J., Yuan, J., Hu, R., Peng, X., Chen, A., Shen, X., 2024. Lightning-generated Whistlers recognition for accurate disaster monitoring in China and its surrounding areas based on a homologous dual-feature information enhancement framework. Remote Sens. Environ. 304, 114021. https://doi.org/10.1016/j.rse.2024.114021

Xiong, S., Zhang, X., Lei, Y., Tan, G., Wang, H., Du, S., 2024. Time-series China urban land use mapping (2016–2022): An approach for achieving spatial-consistency and semantic-transition rationality in temporal domain. Remote Sens. Environ. 312, 114344. https://doi.org/10.1016/j.rse.2024.114344

Xu, L., Herold, M., Tsendbazar, N.-E., Masiliūnas, D., Li, L., Lesiv, M., Fritz, S., Verbesselt, J., 2022. Time series analysis for global land cover change monitoring: A comparison across sensors. Remote Sens. Environ. 271, 112905. https://doi.org/10.1016/j.rse.2022.112905

Yang, B., Qin, L., Liu, J., Liu, X., 2022. UTRNet: An Unsupervised Time-Distance-Guided Convolutional Recurrent Network for Change Detection in Irregularly Collected Images. IEEE Trans. Geosci. Remote Sens. 60, 1–16. https://doi.org/10.1109/TGRS.2022.3174009

Yang, Z., Liu, L., Wang, J., Miao, H., Zhang, Q., 2023. Extrapolation of electromagnetic pointing error corrections for Sentinel-1 Doppler currents from land areas to the open ocean. Remote Sens. Environ. 297, 113788. https://doi.org/10.1016/j.rse.2023.113788

Zhan, W., Wang, C., Wang, S., Li, L., Ji, Y., Du, H., Huang, F., Jiang, S., Liu, Z., Fu, H., 2024. Fraction-dependent variations in cooling efficiency of urban trees across global cities. ISPRS J. Photogramm. Remote Sens. 216, 229–239. https://doi.org/10.1016/j.isprsjprs.2024.07.026

Zhang, C., Chen, Z., Luo, L., Zhu, Q., Fu, Y., Gao, B., Hu, J., Cheng, L., Lv, Q., Yang, J., Li, M., Zhou, L., Wang, Q., 2024. Mapping urban construction sites in China through geospatial data fusion: Methods and applications. Remote Sens. Environ. 315, 114441. https://doi.org/10.1016/j.rse.2024.114441

Zhang, M., Li, D., Li, G., Lu, D., 2024. Vegetation classification in a subtropical region with Sentinel-2 time series data and deep learning. Geo-Spat. Inf. Sci. 1–19. https://doi.org/10.1080/10095020.2024.2336604

Zhao, Z., Ru, L., Wu, C., 2024. Exploring Effective Priors and Efficient Models for Weakly-Supervised Change Detection. https://doi.org/10.48550/arXiv.2307.10853

Zheng, Z., Zhong, Y., Zhang, L., Burke, M., Lobell, D.B., Ermon, S., 2024. Towards transferable building damage assessment via unsupervised single-temporal change adaptation. Remote Sens. Environ. 315, 114416. https://doi.org/10.1016/j.rse.2024.114416

Zhong, H., Wu, C., 2024. T-UNet: triplet UNet for change detection in high-resolution remote sensing images. Geo-Spat. Inf. Sci. 1–18. https://doi.org/10.1080/10095020.2024.2338224

Zhu, Z., Woodcock, C.E., 2014. Continuous change detection and classification of land cover using all available Landsat data. Remote Sens. Environ. 144, 152–171. https://doi.org/10.1016/j.rse.2014.01.011